\documentclass{article} %
\usepackage{iclr2024_conference,times}

\usepackage{times}
\usepackage{epsfig}
\usepackage{graphicx}
\usepackage{amsmath}
\usepackage{wrapfig,lipsum,booktabs}

\usepackage{multirow}
\usepackage{graphicx}  
\usepackage{amssymb}
\usepackage{subfigure}
\usepackage{color}

\usepackage{colortbl}
\usepackage{comment}

\usepackage{hyperref}
\usepackage{url}
\makeatletter
\def\hlinewd#1{%
\noalign{\ifnum0=`}\fi\hrule \@height #1 \futurelet
\reserved@a\@xhline}

\usepackage{pifont}
\definecolor{recolor}{rgb}{0,1,0}

\definecolor{srcolor}{rgb}{1,0,0}

\definecolor{shcolor}{rgb}{0,0,1}

\newcommand{\hlrow}{\rowcolor{black!6}}

\title{Unifying Feature and Cost Aggregation with Transformers for Semantic and Visual Correspondence }

\author{Sunghwan Hong\thanks{Equal contribution} \   , Seokju Cho$^*$, Seungryong Kim\thanks{Corresponding author} \\
Korea University\\
\texttt{\{sung$\_$hwan,seokju$\_$cho,seungryong$\_$kim\}@korea.ac.kr} \\
\And
Stephen Lin \\
Microsoft Research Asia\\
\texttt{stevelin@microsoft.com} \\
}

\iclrfinalcopy %
\begin{document}

\maketitle
\begin{abstract}
    This paper introduces a Transformer-based integrative feature and cost aggregation network designed for dense matching tasks. In the context of dense matching, many works benefit from one of two forms of aggregation: feature aggregation, which pertains to the alignment of similar features, or cost aggregation, a procedure aimed at instilling coherence in the flow estimates across neighboring pixels. In this work, we first show that feature aggregation and cost aggregation exhibit distinct characteristics and reveal the potential for substantial benefits stemming from the judicious use of both aggregation processes. We then introduce a simple yet effective architecture that harnesses self- and cross-attention mechanisms to show that our approach unifies feature aggregation and cost aggregation and effectively harnesses the strengths of both techniques. Within the proposed attention layers, the features and cost volume both complement each other, and the attention layers are interleaved through a coarse-to-fine design to further promote accurate correspondence estimation. Finally at inference, our network produces multi-scale predictions, computes their confidence scores, and selects the most confident flow for final prediction. Our framework is evaluated on standard benchmarks for semantic matching, and also applied to geometric matching, where we show that our approach achieves significant improvements compared to existing methods.
\end{abstract}

\section{Introduction}
Finding visual correspondences between images is a central problem in computer vision, with numerous applications including simultaneous localization and mapping (SLAM)~\citep{bailey2006simultaneous}, augmented reality (AR)~\citep{peebles2021gan}, and structure from motion (SfM)~\citep{schonberger2016structure}. Given visually or semantically similar images, sparse correspondence approaches~\citep{lowe2004distinctive} first detect a set of sparse points and extract corresponding descriptors to find matches across them. In contrast, dense correspondence~\citep{philbin2007object} aims at finding matches for all pixels. Dense correspondence approaches typically follow the classical matching pipeline of feature extraction and aggregation, cost aggregation, and flow estimation~\citep{scharstein2002taxonomy,philbin2007object}.

\begin{figure*}
  \renewcommand{\thesubfigure}{}
\subfigure[(a) Feature Aggregation]
{\includegraphics[width=0.35\linewidth]{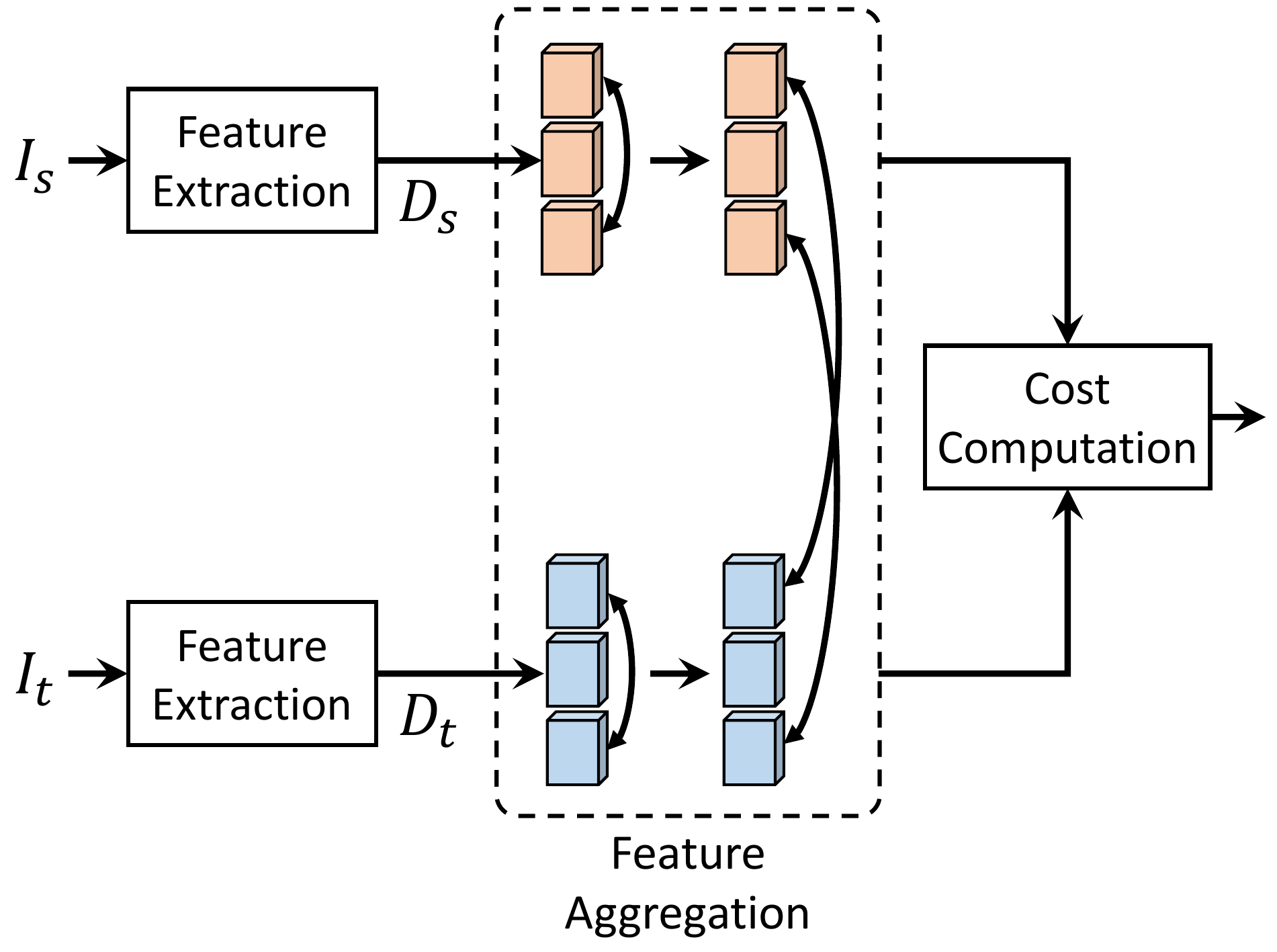}}\hfill
    \subfigure[(b) Cost Aggregation]
{\includegraphics[width=0.25\linewidth]{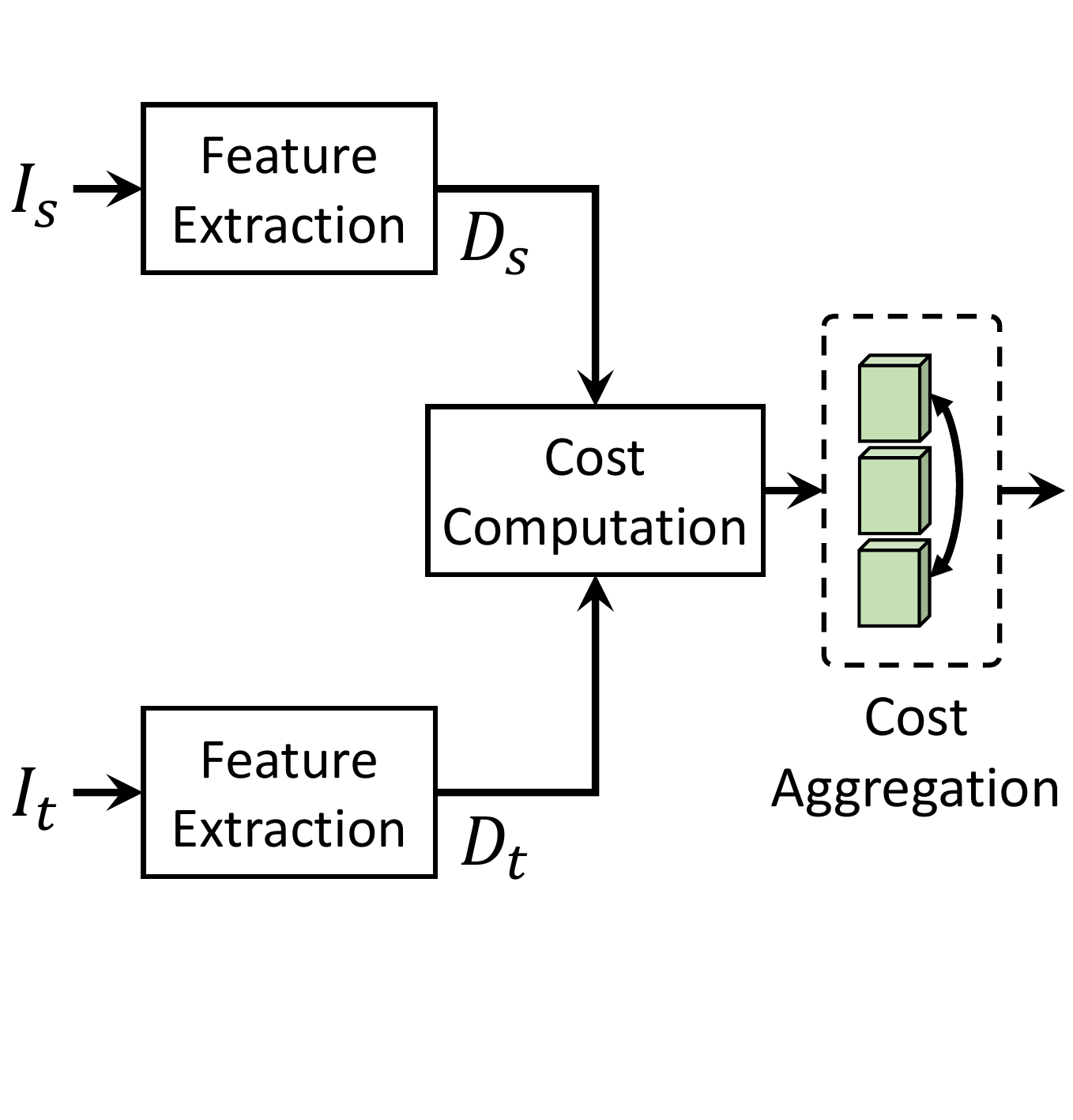}}\hfill
    \subfigure[(c) Integrative Aggregation (Ours)]
{\includegraphics[width=0.377\linewidth]{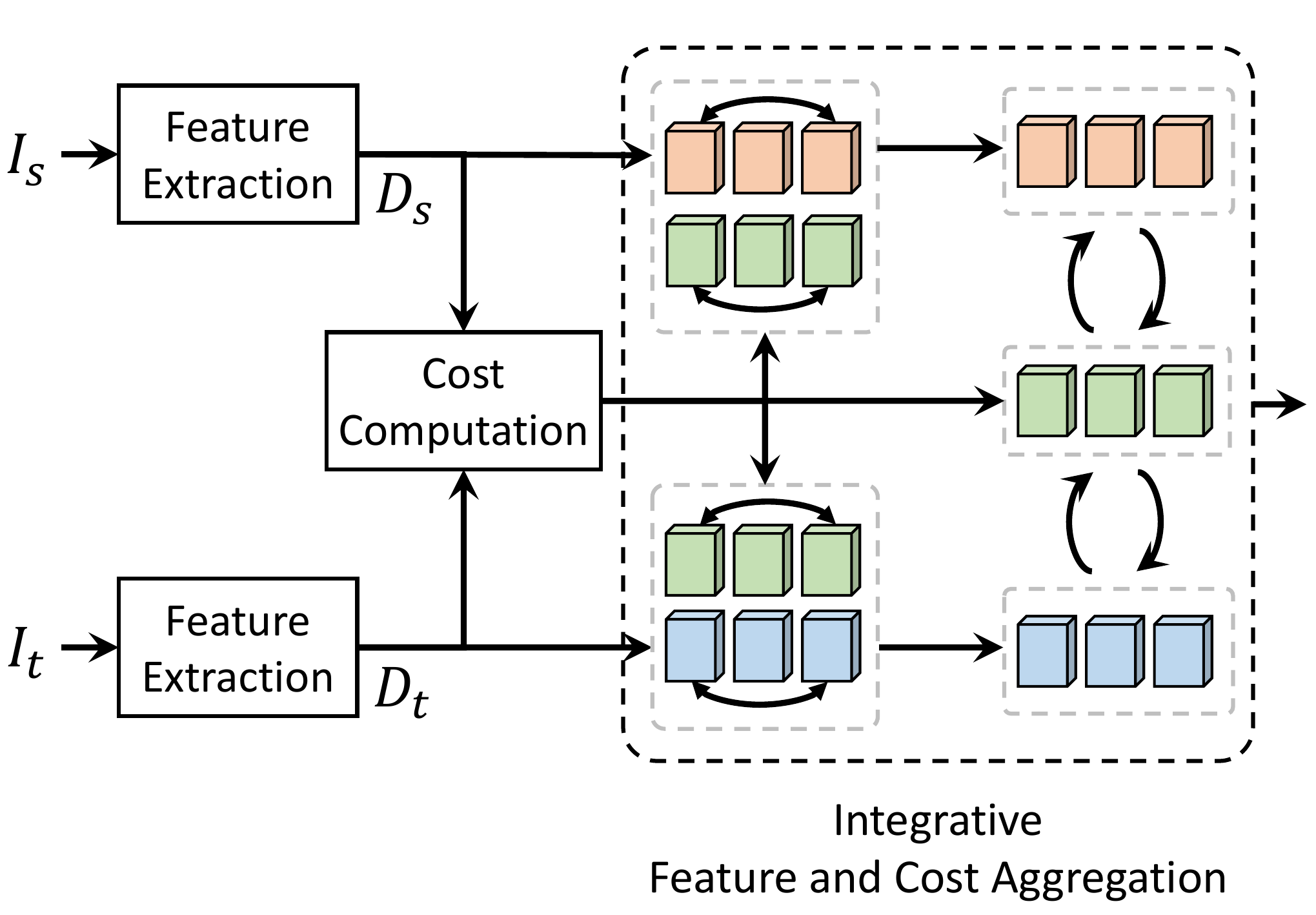}}\hfill\vspace{-10pt}

  \caption{\textbf{Intuition of the proposed method:} (a) feature aggregation methods that aggregate feature descriptors, (b) cost aggregation methods that aggregate a cost volume, and (c) our integrative feature and cost aggregation method, which jointly performs both aggregations to find highly accurate correspondences. }
  \label{intuition}\vspace{-10pt}
\end{figure*}

Much recent correspondence research~\citep{sarlin2020superglue,sun2021loftr,jiang2021cotr,xu2021gmflow,li2021revisiting,cho2021semantic,min2021convolutional,huang2022flowformer,9933865} have utilized a means to benefit from either feature aggregation or cost aggregation. Feature aggregation, as illustrated in Fig.~\ref{intuition} (a), is a process that aims to not only integrate self-similar features within an image but also align similar features between the two images for matching. The advantages of feature aggregation have been made particularly evident in several attention- and Transformer-based matching networks~\citep{vaswani2017attention,sarlin2020superglue,sun2021loftr,xu2021gmflow,jiang2021cotr}. Their accuracy in matching can be attributed to, as we show in Fig.~\ref{PCA} (e-f) and supported by previous studies~\citep{sun2021loftr,amir2021deep}, the learned position-dependent semantic features. While the visualization exhibits consistency among parts with the same semantics, dense matching often requires features with even greater discriminative power for more robust pixel-wise correspondence estimation, which is typically challenged by repetitive patterns and background clutters. 

To compensate for, on the other hand, cost aggregation, as illustrated in Fig.~\ref{intuition} (b), has been adopted by numerous works~\citep{Rocco18b,min2021convolutional++,cho2021semantic,huang2022flowformer,9933865} for its favorable generalization ability~\citep{song2021adastereo,liu2022graftnet} and robustness to repetitive patterns and background clutter, which can be attributed to the matching similarities encoded in the cost volumes. These works can leverage the matching similarities to enforce smoothness and coherence in the disparity or flow estimates across neighboring pixels. However, it is important to note that, as highlighted in Fig.~\ref{fig:multi-cost} (c-h), cost volumes often lack semantic context and exhibit relatively less consideration of spatial structure. This disparity arises due to the fact that the information encapsulated within cost volumes is established on the basis of pixel pairs, which could potentially lead to challenges in scenarios where such contextual cues play a pivotal role.

In this paper, we tackle the dense correspondence task by first performing a thorough exploration of feature aggregation and cost aggregation and their distinct characteristics. We then propose a simple yet effective architecture that can benefit from the potential advantages stemming from a more judicious use of both aggregation processes. The proposed architecture is a Transformer-based aggregation network, namely Unified Feature and Cost Aggregation Transformers (UFC), that models an integrative aggregation of feature descriptors and the cost volume, as illustrated in Fig.~\ref{intuition} (c).

This network consists of two stages, the first of which employs a self-attention layer to aggregate the descriptors and cost volume jointly. In this stage, the descriptors can help to disambiguate the noisy cost volume similarly to cost volume filtering~\citep{hosni2012fast,sun2018pwc}, and the cost volume can encourage the features to account for matching probabilities as an additional factor for alignment.   For the subsequent step, we design a cross-attention layer that enables further aggregation aided by the outputs from earlier aggregations. This aggregated cost volume can guide the alignment with its sharpened matching distribution. These attention layers are interleaved. We further propose hierarchical processing to enhance the benefits one aggregation gains from the other. Finally, at inference time, our method estimates multi-scale predictions and their confidence scores to recover highly accurate flow. %

We first evaluate the proposed method on the tasks of semantic matching and subsequently, we also substantiate that our framework achieves appreciable performance when applied to geometric matching. Our framework clearly outperforms prior works on all the major dense matching benchmarks, including   HPatches~\citep{balntas2017hpatches}, ETH3D~\citep{schops2017multi}, SPair-71k~\citep{min2019spair}, PF-PASCAL~\citep{ham2017proposal} and PF-WILLOW~\citep{ham2016proposal}. We provide extensive ablation and analysis to validate our design choices. %

\section{Related Work}
\paragraph{Feature Extraction and Aggregation.}
Feature extraction involves detecting interest points and extracting the descriptors of the corresponding points. In traditional methods~\citep{liu2010sift,bay2006surf,dalal2005histograms,tola2009daisy}, the matching performance mostly relies on the quality of the feature detection and description methods, and outlier rejection across matched points is typically determined by RANSAC~\citep{fischler1981random}.%

Learning-based feature extraction methods~\citep{detone2018superpoint,ono2018lf,dusmanu2019d2,revaud2019r2d2} obtain dense deep features tailored for matching. These works have demonstrated that the quality of feature descriptors contributes substantially to matching performance. In accordance with this, recent matching networks~\citep{sarlin2020superglue,min2019hyperpixel,lee2019sfnet,Hong_2021_ICCV,min2020learning,jiang2021cotr,sun2021loftr,xu2021gmflow} proposed effective means for feature aggregation. 
Notable sparse correspondence works include SuperGlue~\citep{sarlin2020superglue} and LOFTR~\citep{sun2021loftr}, which employ graph- or Transformer-based self- and cross-attention for aggregation. Other methods are PUMP~\citep{revaud2022pump} and ECO-TR~\citep{tan2022eco}, which are follow-up works of COTR~\citep{jiang2021cotr}. These methods evaluate on dense correspondence benchmarks in a different way from previous works, using only the sparse and quasi-dense correspondences above certain confidence levels. We refer the readers to the supplementary material for a detailed discussion.

For dense correspondence, SFNet~\citep{lee2019sfnet} and DMP~\citep{Hong_2021_ICCV} introduce adaptation layers after feature extraction to learn feature maps well-suited to matching and are evaluated on dense semantic and geometric matching. DKM~\citep{edstedt2023dkm} adopts Gaussian Processing Kernels for dense correspondence, and it demonstrates its effectiveness in pose estimation. In optical flow,   GMFlow~\citep{xu2021gmflow} leverages Transformer for feature aggregation, and its extension~\cite{xu2023unifying} applies the method to stereo matching and depth estimation. In semantic correspondence, notable works include SCorrSAN~\citep{huang2022learning} proposes an efficient spatial context encoder to aggregate spatial context and feature descriptors, and MMNet~\cite{zhao2021multi} proposes a multi-scale matching network to learn discriminative pixel-level features. 

\vspace{-10pt}

\paragraph{Cost Aggregation.}
In the dense correspondence literature, many works have designed their architectures for effective cost aggregation, which brings strong generalization power~\citep{song2021adastereo,liu2022graftnet}. Recent works~\citep{truong2020glu,Hong_2021_ICCV,jeon2020guided,truong2021learning} use 2D convolutions to establish correspondence while aggregating the cost volume with learnable kernels,  while some works~\citep{min2019hyperpixel,min2020learning,liu2020semantic} utilize handcrafted methods, which include RHM~\citep{cho2015unsupervised} and the OT solver~\citep{sinkhorn1967diagonal}. NC-Net~\citep{Rocco18b} was the first to propose 4D convolutions for cost aggregation, and numerous works~\citep{li2020correspondence,yang2019volumetric,huang2019dynamic,rocco2020efficient,min2021convolutional,min2021convolutional++} leveraged or extended 4D convolutions. 

Among Transformer-based networks, CATs~\citep{cho2021semantic} recently proposed to use Transformer~\citep{vaswani2017attention} for cost aggregation, and its extension CATs++~\citep{9933865} combined convolutions and Transformer for an enhanced cost aggregation.  VAT~\citep{hong2022cost} proposed 4D convolutional Swin transformer for cost aggregation that benefits from  better generalization power and showed its effectiveness for semantic correspondence. NeMF~\citep{hong2022neural} incorporates an implicit neural representation into semantic correspondence and implicitly represents the cost volume to infer correspondences defined at arbitrary resolution. FlowFormer~\citep{huang2022flowformer} and STTR~\citep{li2021revisiting} are Transformer-based cost aggregation networks specifically designed for optical flow or stereo matching.

\section{Feature and Cost Aggregation}\label{motivation}
\begin{figure*}
  \centering
  \renewcommand{\thesubfigure}{}
    \subfigure[ ]
{\includegraphics[width=0.12\linewidth]{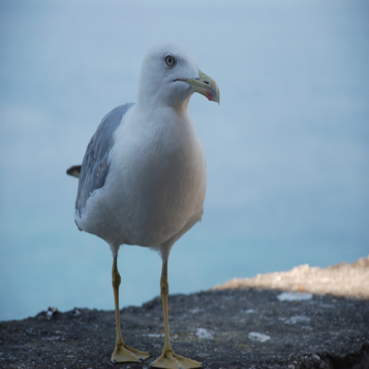}}\hfill
    \subfigure[ ]
{\includegraphics[width=0.12\linewidth]{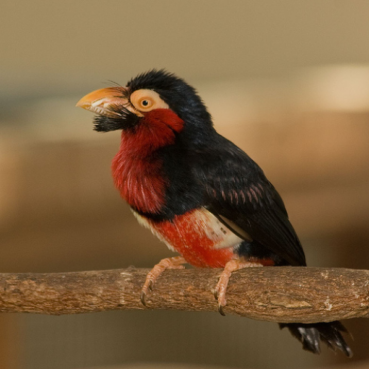}}\hfill
    \subfigure[]
{\includegraphics[width=0.12\linewidth]{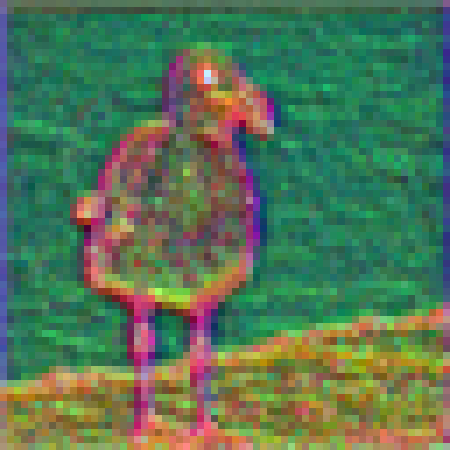}}\hfill
    \subfigure[]
{\includegraphics[width=0.12\linewidth]{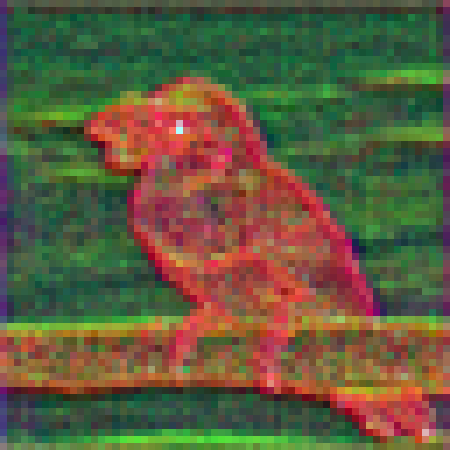}}\hfill
    \subfigure[  ]
{\includegraphics[width=0.12\linewidth]{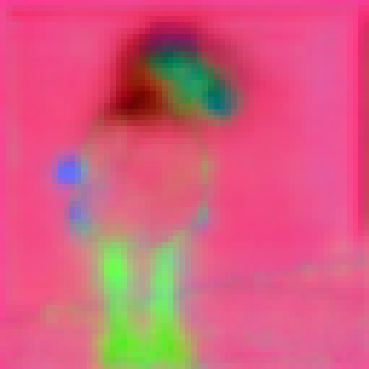}}\hfill
    \subfigure[ ]
{\includegraphics[width=0.12\linewidth]{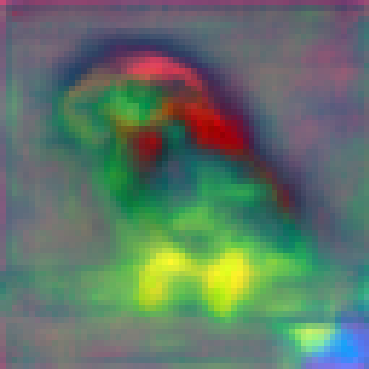}}\hfill
    \subfigure[]
{\includegraphics[width=0.12\linewidth]{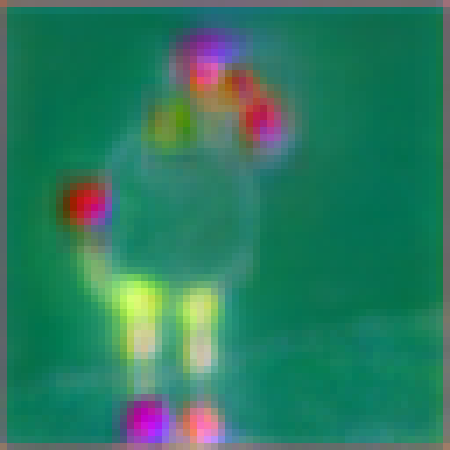}}\hfill
    \subfigure[]
{\includegraphics[width=0.12\linewidth]{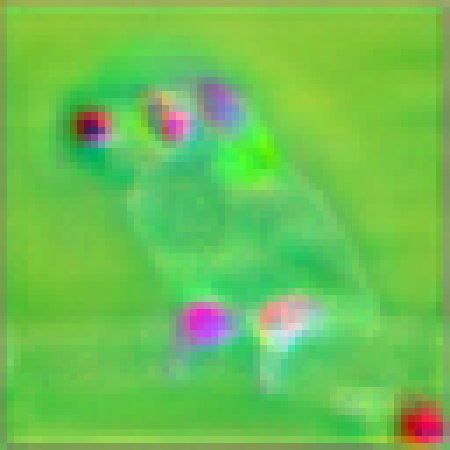}}\hfill
\\\vspace{-20pt}
    \subfigure[(a) ]
{\includegraphics[width=0.12\linewidth]{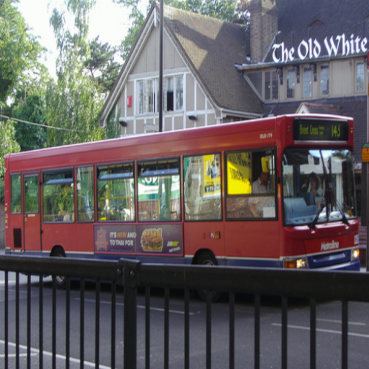}}\hfill
    \subfigure[(b) ]
{\includegraphics[width=0.12\linewidth]{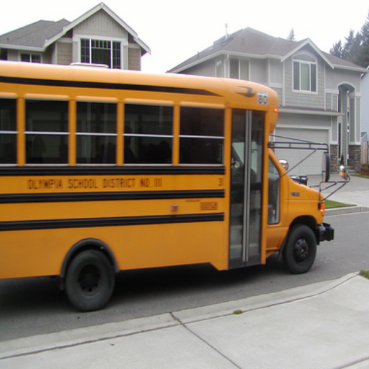}}\hfill
    \subfigure[(c) ]
{\includegraphics[width=0.12\linewidth]{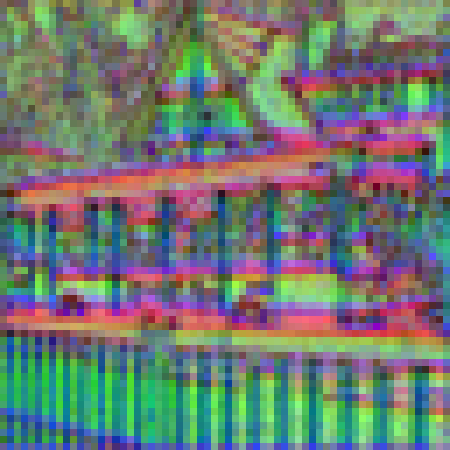}}\hfill
    \subfigure[(d)]
{\includegraphics[width=0.12\linewidth]{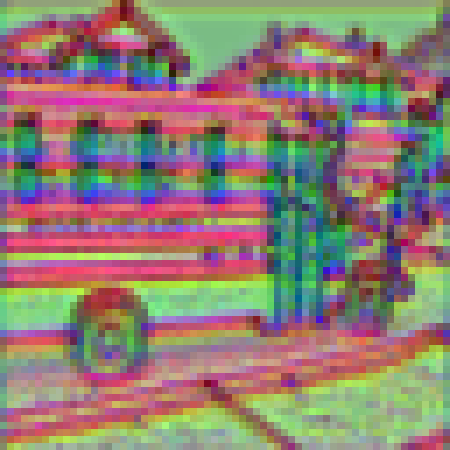}}\hfill
    \subfigure[(e)  ]
{\includegraphics[width=0.12\linewidth]{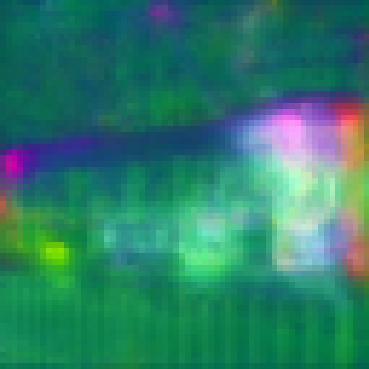}}\hfill
    \subfigure[(f) ]
{\includegraphics[width=0.12\linewidth]{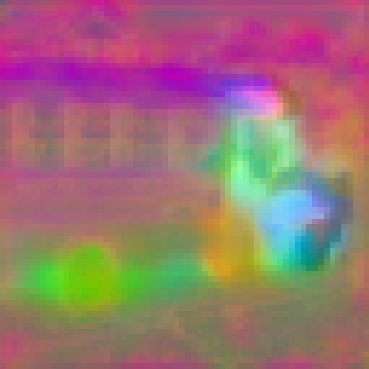}}\hfill
    \subfigure[(g)]
{\includegraphics[width=0.12\linewidth]{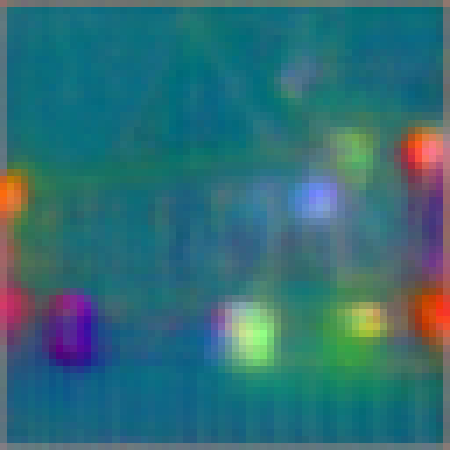}}\hfill
    \subfigure[(h) ]
{\includegraphics[width=0.12\linewidth]{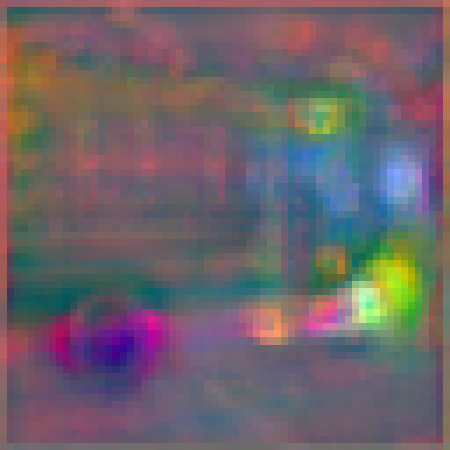}}\hfill
\\\vspace{-10pt}
  \caption{\textbf{\textbf{PCA} visualizations of feature maps:} (a-b) source and target images. (c-d) raw feature maps. (e-f) feature maps that have undergone feature aggregation.  (g-h) feature maps that have undergone  integrative aggregation. Our integrative aggregation methodology enables the acquisition of more discerning feature representations while preserving both semantic and spatial structural aspects, resulting in the estimation of highly accurate correspondences. }
  \label{PCA}\vspace{-10pt}
\end{figure*}

\begin{figure*}
  \centering
  \renewcommand{\thesubfigure}{}
    \subfigure[]
{\includegraphics[width=0.12\linewidth]{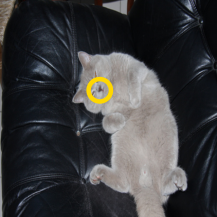}}\hfill
    \subfigure[]
{\includegraphics[width=0.12\linewidth]{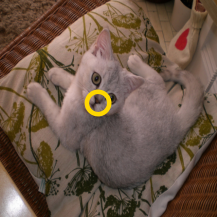}}\hfill
    \subfigure[]
{\includegraphics[width=0.12\linewidth]{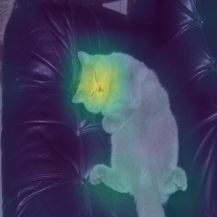}}\hfill
    \subfigure[]
{\includegraphics[width=0.12\linewidth]{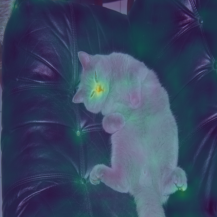}}\hfill
    \subfigure[]
{\includegraphics[width=0.12\linewidth]{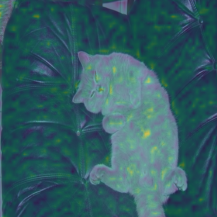}}\hfill
    \subfigure[]
{\includegraphics[width=0.12\linewidth]{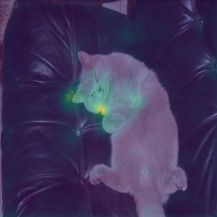}}\hfill
    \subfigure[]
{\includegraphics[width=0.12\linewidth]{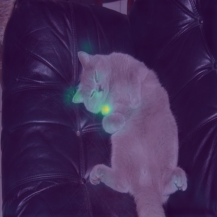}}\hfill
    \subfigure[]
{\includegraphics[width=0.12\linewidth]{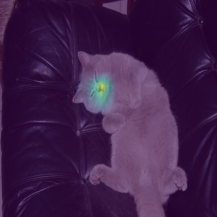}}\hfill
\\\vspace{-20pt}
    \subfigure[(a)]
{\includegraphics[width=0.12\linewidth]{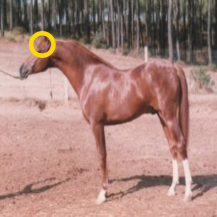}}\hfill
    \subfigure[(b)]
{\includegraphics[width=0.12\linewidth]{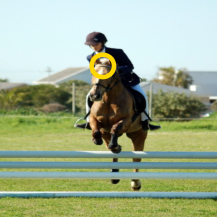}}\hfill
    \subfigure[(c)]
{\includegraphics[width=0.12\linewidth]{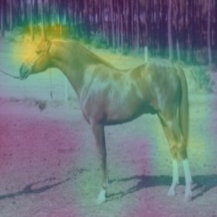}}\hfill
    \subfigure[(d)]
{\includegraphics[width=0.12\linewidth]{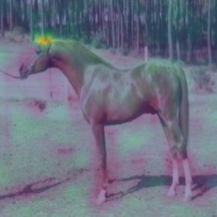}}\hfill
    \subfigure[(e)]
{\includegraphics[width=0.12\linewidth]{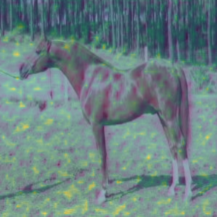}}\hfill
    \subfigure[(f)]
{\includegraphics[width=0.12\linewidth]{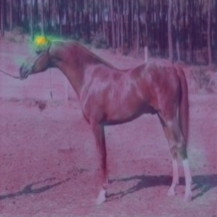}}\hfill
    \subfigure[(g)]
{\includegraphics[width=0.12\linewidth]{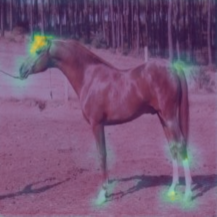}}\hfill
    \subfigure[(h)]
{\includegraphics[width=0.12\linewidth]{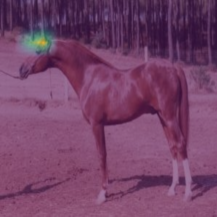}}\hfill
\\\vspace{-10pt}

  \caption{\textbf{Visualizations of cost volumes:} (a-b) source and target images. (c-e) \textbf{2D slices} of raw cost volumes at different levels $l$. (f) cost volumes constructed using feature maps that have undergone self-attention. (g) with feature maps that have undergone both self-attention and cross-attention, and (h) the cost volume that have undergone cost aggregation. From (c-e), the noises are suppressed in (h), while (f-g) shows aggregated features help to construct less noisy cost volumes. Note that the visualizations are obtained with respect to the circled point in the target image.}
  \label{fig:multi-cost}\vspace{-10pt}
\end{figure*}

In this section, we examine the characteristics of feature and cost aggregation, which will later be verified empirically in Section~\ref{quan}. Fig.~\ref{PCA} and Fig.~\ref{fig:multi-cost} present visualizations of feature maps and cost volumes at different stages of aggregation. Although there may be different types of aggregations, throughout this work, we focus on attention-based aggregations. From the visualizations, the following observations can be made. 

\textbf{The information encoded by features and cost volumes differ.} Feature aggregation and cost aggregation thus exploit different information, which is exemplified in Fig.~\ref{PCA} (c-d) and Fig.~\ref{fig:multi-cost} (c-e) where the spatial structure is preserved in the features while sparse spatial locations with higher similarity to the query point is highlighted in the cost volume. Due to the different information encoded in their inputs, \textbf{the outputs of both aggregations have different characteristics.} In Fig.~\ref{PCA}, compared to raw features in (c-d), the feature aggregation in (e-f) makes the features of semantic parts, \textit{e.g.,} legs/claws and head, more consistent between the two birds. This is in agreement with observations in previous studies~\citep{amir2021deep,sun2021loftr}. On the other hand, compared to noisy cost volumes visualized in Fig.~\ref{fig:multi-cost} (c-e), the aggregated cost volume in (h) is less noisy and more clearly highlights the most probable region for matching while less probable regions are suppressed. We additionally observe that \textbf{each type of aggregation can have apparent effects on the other.} Naturally, more robust descriptors can construct a less noisy cost volume as shown in Fig.~\ref{fig:multi-cost} (f-g), and ease the subsequent cost aggregation process to promote more accurate correspondences. However, without a special model design, cost aggregation would not affect feature aggregation since it is performed after feature aggregation.  Motivated by this, we  proposed integrative aggregation, and its effects on feature maps are shown in Fig.~\ref{PCA} (g-h), where salient features become more distinguishable from each other, \textit{i.e.,} the left and right claws. 
This exemplifies that more discriminative feature representations that are more focused  and that also preserve semantics and spatial structure can be learned and reveals that potential benefits from their integration can be realized.

From these observations, we propose in the following a simple yet effective architecture that makes prudent use of both feature and cost aggregation.

\section{Methodology}\label{sec:3}

\subsection{Problem Formulation}
Let us first denote a pair of visually or semantically similar images, i.e., the source and target, as $I_{s}$ and $I_{t}$, the feature descriptors extracted from $I_{s}$ and $I_{t}$ as $D_{s}$ and $D_{t}$, respectively, and the cost volume computed between the feature maps as $C$. Given $I_s$ and $I_t$, we aim to establish a dense correspondence field ${F}(i)$ that is defined at all pixels $i$ and warps $I_s$ towards $I_t$. Given features extracted from deep CNNs~\citep{he2016deep} or Transformers~\citep{dosovitskiy2020image}, we can construct and store a cost volume that consists of all pairwise feature similarities $C \in \mathbb{R}^{h \times w \times h \times w}$ with height $h$ and width $w$: $C(i,j)=D_{s}(i)\cdot {D}_{t}(j)$,
where $i$ and $j$ index the source and target features, respectively. The dense correspondence field, ${F}(i)$, can then be determined from $C(i,j)$ considering all $j$.
\subsection{Preliminaries: Self- and cross-attention}
We briefly explain the attention mechanism, a core component we extend from. Given a sequence of tokens as an input, Transformer~\citep{vaswani2017attention} first linearly projects tokens to obtain query, key and value embeddings. These are then fed into a scaled dot product attention layer, followed by Layer Normalization (LN)~\citep{ba2016layer} and a feed-forward network or MLP, to produce an output with the same shape as the input. Each token is attended to by all the other tokens. This projections are formulated as:
\begin{equation}
\begin{split}
        Q = \mathcal{P}_Q(X),\quad
        K = \mathcal{P}_K(X), \quad
        V = \mathcal{P}_V(X),
\end{split}
\end{equation}
where $\mathcal{P}_Q$, $\mathcal{P}_K$ and $\mathcal{P}_V$ denote query, key and value projections, respectively, and $X$ denotes a token with a positional embedding. Subsequently, they pass through an attention layer: 
\begin{equation}
    \mathrm{Attention}(X) = \mathrm{softmax}(\frac{QK^T}{\sqrt{d_K}})V,
    \label{eq:2}
\end{equation}
where $d_K$ is the dimension of the key embedding. Note that the $\mathrm{Attention}(\cdot)$ function can be defined in various ways~\citep{wang2020linformer,liu2021swin,katharopoulos2020transformers,lu2021soft,wu2021fastformer}. Self- and cross-attention are distinguished by their input to the key and value projections. Given a pair of input tokens, e.g., $X_s$ and $X_t$, the input to the key and value projections of self-attention for  $X_s$ is the same input, $X_s$, but for cross-attention across $X_s$ and $X_t$, the inputs to the key and value projection are $X_t$.

\begin{figure*}
 \renewcommand{\thesubfigure}{}
\subfigure[(a) Integrative self-attention]
{\includegraphics[width=0.57\linewidth]{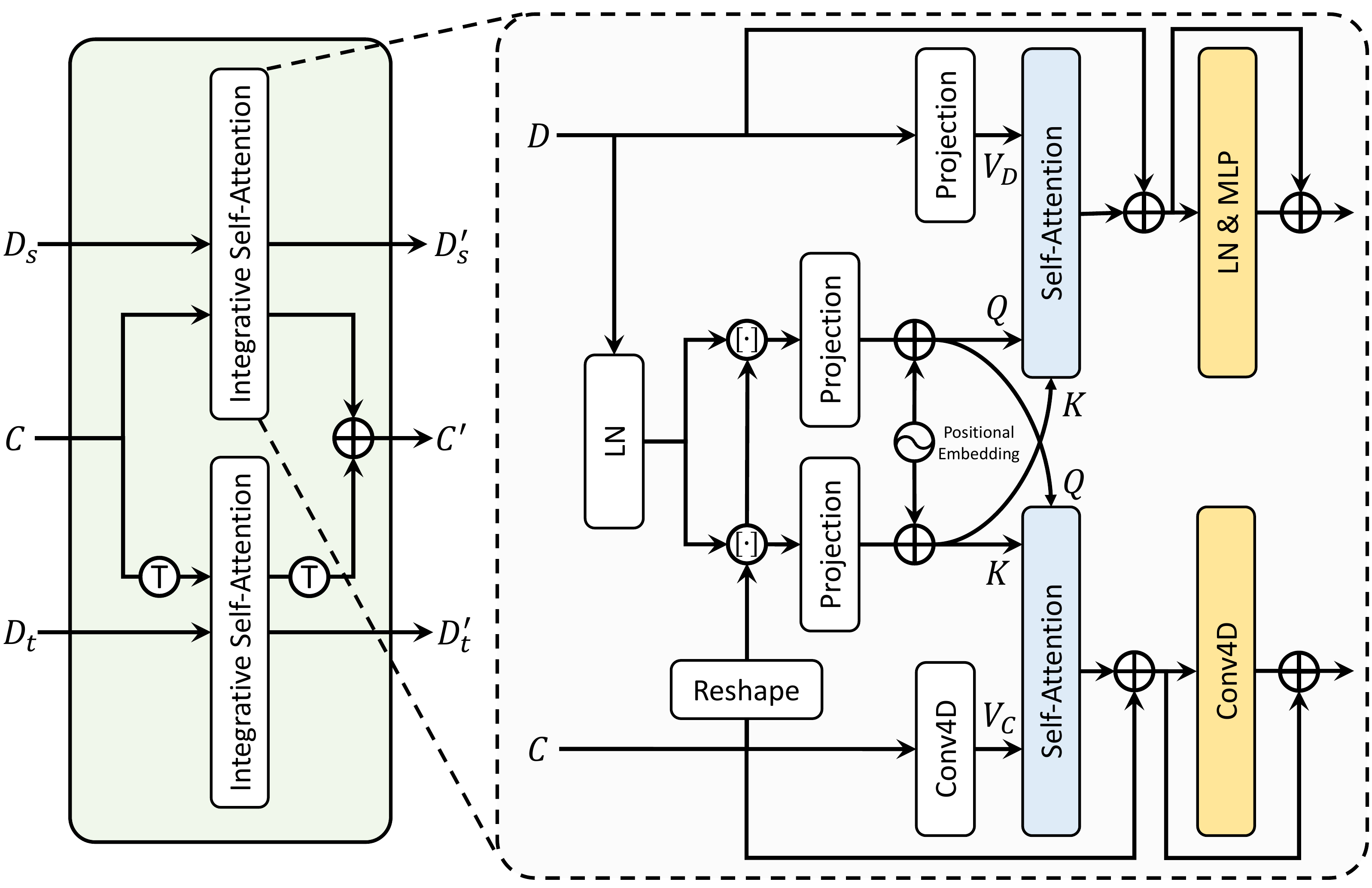}}\hfill
    \subfigure[(b) Cross-attention with matching distribution]
{\includegraphics[width=0.40\linewidth]{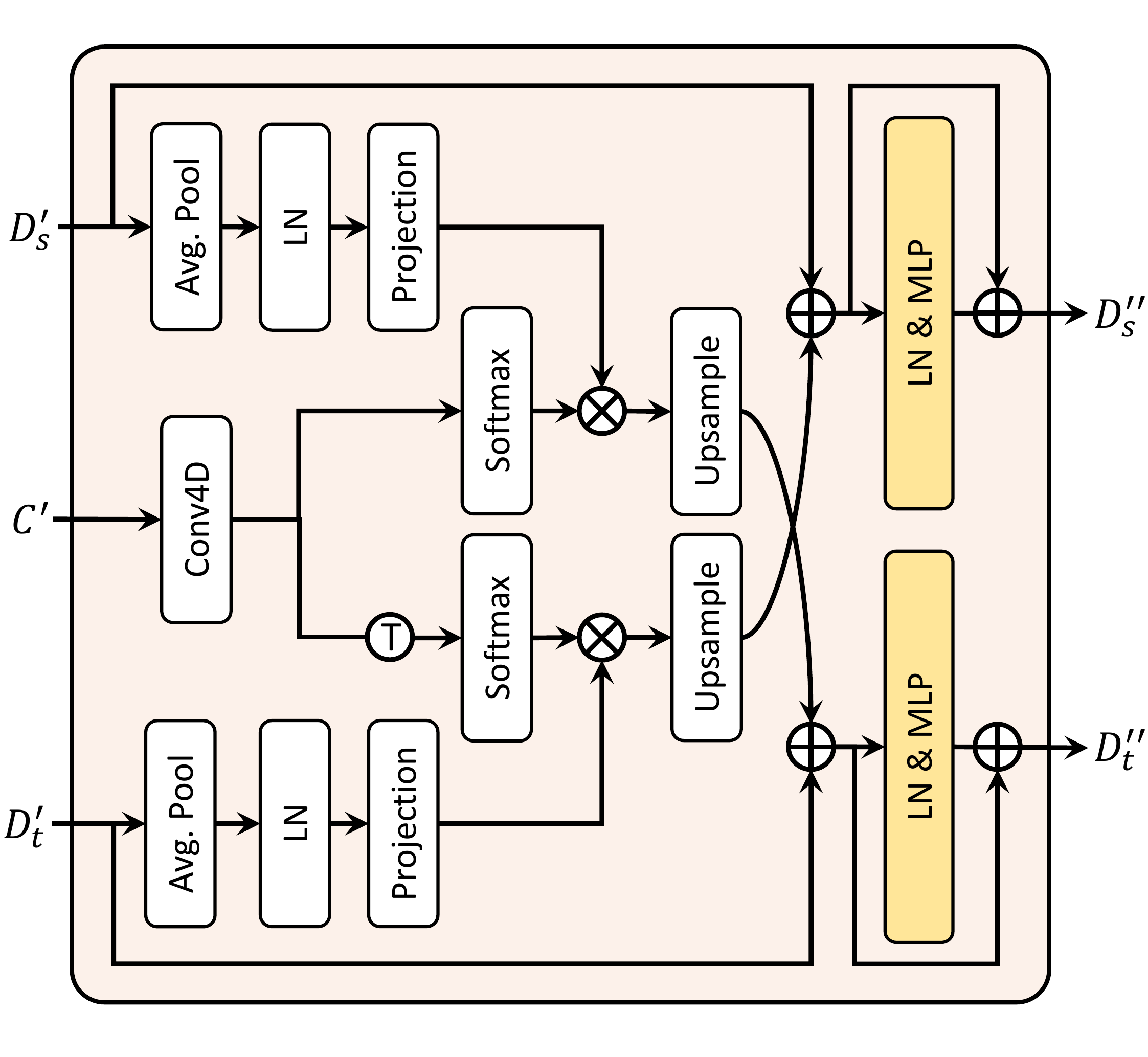}}\hfill

  \caption{\textbf{Illustration of the proposed self- and cross-attention:} (a) joint feature aggregation and cost aggregation, and (b) cross-attention layer with matching distribution. }
  \label{attention}\vspace{-10pt}
\end{figure*}

\subsection{Unified Feature and Cost Aggregation}

\paragraph{Integrative Self-Attention.}

Toward more judicious use of both aggregations, we first leverage the fact that  both feature descriptors $D_s,$ $D_t$ and cost volume $C$ encode different information. To this end, in the proposed integrative self-attention layer, as shown in Fig.~\ref{attention}, we first obtain a feature cost volume $[D,C]$ by concatenating $D$ and $C$, where $[\cdot,\cdot]$ denotes concatenation. 

This concatenation brings benefits from two perspectives. From the cost aggregation point of view, the feature map of the feature cost volume can disambiguate the initial noisy cost volume by referring to semantic-aware features as demonstrated in the stereo matching literature~\citep{yoon2006adaptive,hosni2012fast,he2011global},~\textit{i.e.,} 
cost volume filtering. From the feature aggregation point of view, the cost volume explicitly represents the similarity of features in one image with respect to the features in the other, and accounting for it drives the features in each image to become more compatible with those of the other. As the iterations unfold, this process will encourage both aggregations to benefit each other. In the end, the resultant feature representations will be more robust and discriminative as shown in Fig.~\ref{PCA} (g-h).   

To compute self-attention, we take a different approach from other works~\citep{sun2021loftr,cho2021semantic} to define the query, key and value embeddings. Concretely, we define two independent value embeddings, specifically one for feature projection and the other for cost volume projection. Formally, we define the query, key and values as:
\begin{equation}
    \begin{split}
&Q = \mathcal{P}_{Q}([D , C]), \quad K = \mathcal{P}_{K}([D , C]), \\ &V_{D} = \mathcal{P}_{V_D}(D), \quad V_{C} = \mathcal{P}_{V_C}(C),
    \end{split}
\label{eq:cost_volume}
\end{equation}    
where $V_D$ and $V_C$ denote the value embeddings of feature descriptors and the cost volume, respectively. After computing an attention map by applying softmax over the query and key dot product, we use it to aggregate feature $D$ and cost volume $C$ with $V_D$ and $V_C$ using Eq.~\ref{eq:2} as follows:
\begin{equation}
    \begin{split}
    \mathrm{Attention}_\mathrm{self-D}(C,D) = \mathrm{softmax}(\frac{QK^T}{\sqrt{d_K}})V_D, \\
    \mathrm{Attention}_\mathrm{self-C}(C,D) = \mathrm{softmax}(\frac{QK^T}{\sqrt{d_K}})V_C.
    \end{split}
\end{equation}
Note that any type of attention computation can be utilized,~\textit{i.e.,} additive~\citep{bahdanau2014neural} or dot product~\citep{vaswani2017attention}, while in practice we use the linear kernel dot product with the associative property of matrix products~\citep{katharopoulos2020transformers}. The outputs of this self-attention are denoted as $D'_s$, $D'_t$, and $C'$.

\vspace{-10pt}

\paragraph{Cross-Attention with Matching Distribution.}
In the proposed cross-attention layer, the aggregated features and cost volume are explicitly used for further aggregation, and we condition both feature descriptors on both input images via this layer. By exploiting the outputs of the self-attention layer, the cross-attention layer performs cross-attention between feature descriptors for further feature aggregation using the improved feature descriptors $D'_s$, $D'_t$ and enhanced cost volume $C'$ from earlier aggregations. 

As shown in Fig.~\ref{attention}, we first apply convolution to the input cost volume and treat the output as a cross-attention map, since applying a softmax function over the cost volume is tantamount to obtaining an attention map. 
In this way, an enhanced aggregation is enabled, as the input cost volume is transformed to represent a sharpened matching distribution. With a cross-attention map and value for the attention score defined as $QK^T = C'$ and $V_{D'} = \mathcal{P}_{V_D}(D')$, respectively, the subsequent attention process for cross-attention is then defined as follows:
\begin{equation}
    \mathrm{Attention}_\mathrm{cross}(C',D') = \mathrm{softmax}(\frac{C'}{\sqrt{d_K}})V_{D'}.
    \label{2}
\end{equation}
The outputs of this cross-attention are denoted as $D''_s$ and $D''_t$, and $C''$ is constructed using  $D''_s$ and $D''_t$. The proposed attention layers are interleaved, and they are stacked $N$ times to facilitate the aggregations and increase the model capacity.

\begin{figure*}
    \centering
    \includegraphics[width=1\linewidth]{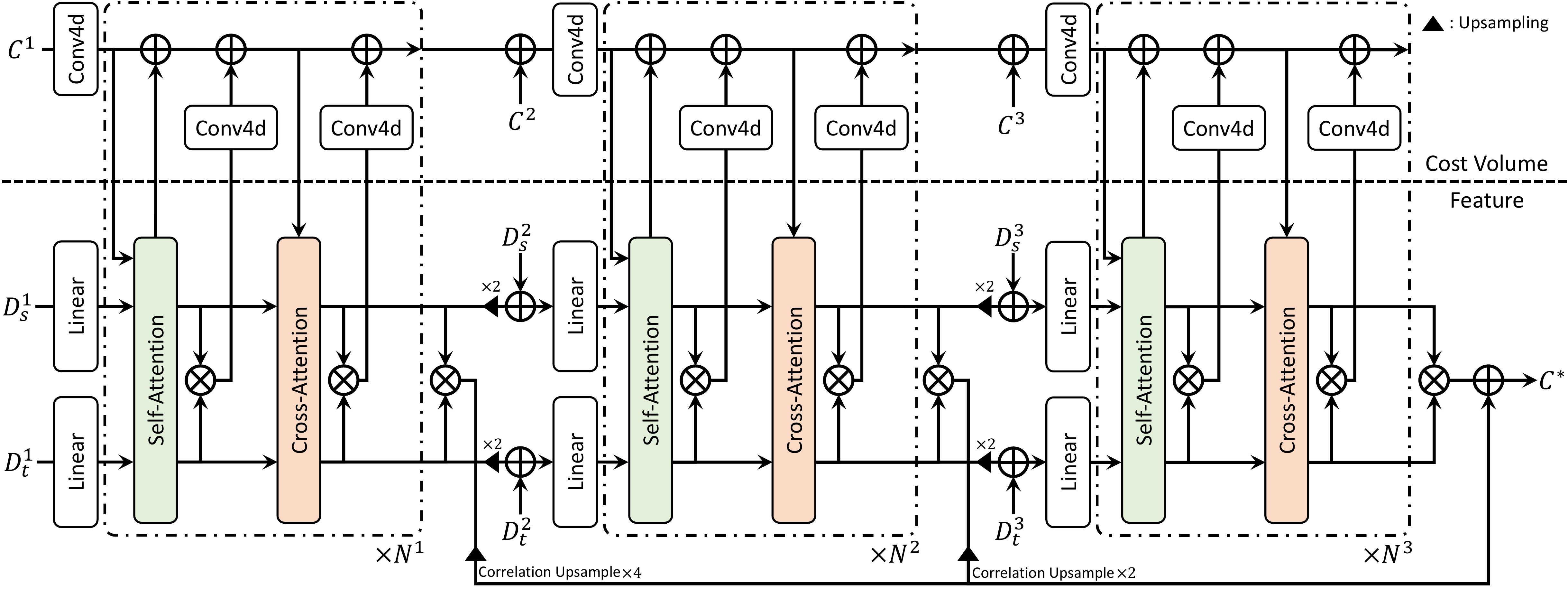}\hfill\\
    \vspace{+5pt}
    \caption{\textbf{Overall architecture of the proposed method.} Given feature maps $D_s$ and $D_t$ and the cost volume $C$ as inputs, our method employs self- and cross-attention specifically designed to conduct joint feature aggregation and cost aggregation in a coarse-to-fine manner.}
    \label{fig:overall}\vspace{-10pt}
\end{figure*}

\subsection{Coarse-to-Fine Formulation }\label{coarse-to-fine}
To improve the robustness of fine-scale estimates and enhance the benefits one aggregation gains from the other, we extend our architecture to a coarse-to-fine approach through pyramidal processing, as done in~\citep{jeon2018parn,melekhov2019dgc,truong2020glu,Hong_2021_ICCV}. 
We first use a coarse pair of refined feature maps and aggregated cost volume, and similar to~\cite{zhao2021multi} that learns complementary correspondence by adding the cost volume of the previous scale, we progressively learn complementary descriptors and correspondences and encourage the coarser outputs to enhance the subsequent aggregations.  

Formally, given the outputs of the attention block at each level, ${D}''^{,l}_s, {D}''^{,l}_t$ and ${C}''^{,l}$, where $l$ denotes the $l$-th level, we upsample the aggregated features using bilinear interpolation and add them to the raw feature descriptors extracted from $I_s$ and $I_t$ defined at the next level: ${D}^{{l+1}}_s = {D}^{{l+1}}_s + \mathrm{up}({D}''^{,l}_s)$, where ${D}^{{l+1}}_t$ is defined similarly.  Note that we let the output cost volumes of self- and cross-attention at each level, ${C}'^{,l}$ and ${C}''^{,l}$, undergo convolution and residual connections, \textit{i.e.,} ${C}''^{,l} = {C}'^{,l} + \mathrm{Conv4d}({C}''^{,l}),$ to facilitate the training process. Then, we define the next-level cost volume as ${C}^{l+1} = {C}^{l+1} + {C}''^{,l} $.

Finally, given the features ${D}''_s$ and ${D}''_t$ at each level, we compute the cost volume, and the sum of all cost volumes across all levels are added up to obtain the final output ${C}^*$ that is used to estimate the final flow field, as shown in the bottom of Fig.~\ref{fig:overall}.
\subsection{Inference: Dense Zoom-in}
At the inference phase, we leverage multi-scale predictions to predict highly accurate correspondences. The goals of this approach are two-fold: to prevent a large memory increase when processing high-resolution input image pairs, \textit{e.g.,} HD or Full HD, and to capture possible fine-grained correspondences missed by the coarse-to-fine design. 

Dense zoom-in consists of three stages. For the first stage, UFC takes an input image pair and uses the output flow to coarsely align the source image to the target image as similarly done in~\citep{shen2020ransac}. Unlike RANSAC-Flow~\citep{shen2020ransac}, we do not resort to finding a homography transformation, but rather rely on the output flow itself. We empirically find that for images with extreme geometric deformations, reliable homography transformations may not be found.

Subsequently, we evenly partition the coarsely aligned source image and the target image into $k \times k $ local windows, where $k$ is a hyperparameter. Each pair of partitioned local windows at the same location is then used to find more fine-grained correspondences by feeding them into UFC to obtain local flow fields. Note that in this stage, we also compute a cycle-consistency confidence score~\citep{jiang2021cotr} that will be used in the final decision-making process. To enable multi-scale inference, we choose multiple $k$, for which we provide an ablation study in the supplementary material.  We then perform transitive composition~\citep{zhou2016learning} using the coarse flow and each of the multi-scale flows. Finally, using the confidence values of composited flows at each pixel, we select the flow with the highest confidence score. This selection is performed for every pixel and results in a final dense flow map.

\section{Experiments}

\begin{table*}[!t]
    \begin{center}

    \scalebox{0.54}{
    \begin{tabular}{l|cc|cc|ccccc|ccc|cc|cc}
            \hlinewd{0.8pt}
             \multirow{3}{*}{Methods}&{Train}&{Keypoint}&\multirow{3}{*}{Feat. Agg.} & \multirow{3}{*}{Cost Agg.}&  \multicolumn{5}{c|}{SPair-71k} & \multicolumn{3}{c|}{PF-PASCAL} & \multicolumn{4}{c}{PF-WILLOW} \\
          
              &Image&Annotation&& &\multicolumn{5}{c|}{PCK @ $\alpha_{\text{bbox}}$} & \multicolumn{3}{c|}{PCK @ $\alpha_{\text{img}}$}  & \multicolumn{2}{c}{PCK @ $\alpha_{\text{bbox}}$} & \multicolumn{2}{c}{PCK @ $\alpha_{\text{bbox-kp}}$} \\ 
        
             &Reso.&Reso.& &&0.01&0.03 &0.05&0.1&0.15 & 0.05 & 0.1 & 0.15 & 0.05 & 0.1 & 0.05 &0.1 \\ 
             \midrule\midrule

             DHPF &240$\times$ 240&240 &-&RHM &1.74&11.0&20.9&37.3 &47.5&75.7 &90.7 &95.0 &49.5 &77.6 &- &71.0\\
             SCOT&-&Max 300 &-&OT-RHM &-&-&-&35.6& -&63.1 &85.4 &92.7 &- &- &{47.8}&\underline{76.0}\\
             CHM&240$\times$ 240&240 & 2D Conv.& 6D Conv. &2.25&14.9&27.2&46.3&57.5 &80.1 &91.6 &94.9 &{52.7} &{79.4} &-&69.6\\

             CATs &256$\times$ 256&256&-&Trans.&1.90&13.8&27.7&49.9&61.7 &75.4 &{92.6} &{96.4} &50.3 &79.2& 40.7&69.0\\
                  MMNet-{FCN} &224$\times$ 320&224$\times$ 320&Conv. + Trans.&-&2.80&18.8  &33.3& 50.4&61.2&81.1 & 91.6& 95.9&- &- &-&-\\
              PWarpC-NC-Net&400$\times$ 400&Ori  &-&4D Conv.&2.55&17.1&31.6&{52.0}&61.8&79.2&92.1&95.6&-&-&\underline{48.0}&\textbf{76.2}\\

             SCorrSAN&256$\times$ 256&256& Linear&-&-&-&-&{55.3} &-&81.5 &{93.3} &{96.6} &{54.1} &{80.0} & -&-\\
               VAT &512$\times$ 512& 512 &-&4D Conv. + Trans.&3.17&19.6&35.0&{55.5} &65.1&78.2 &{92.3} &{96.2} &{52.8} &\textbf{81.6} &42.3 &71.3\\
               CATs++&512$\times$ 512& 512 &-&4D Conv. + Trans.&\underline{4.31}&\underline{25.0}&\underline{40.7}&\underline{59.8} &\underline{68.5}&\underline{84.9} &\underline{93.8} &\underline{96.8} &\underline{56.7} &\underline{81.2} & 47.0&72.6\\
               TransforMatcher&240$\times$ 240& 240 &-&Trans.&-&-&-&53.7 &-&80.8 &91.8 &-&-&76.0 &-&65.3 \\
               NeMF&512$\times$ 512& Ori  &-&4D Conv. + Trans.&3.2&19.5&34.2&53.6 &-&80.6 &93.6 &- &- &- &60.8 &75.0\\
      
          \midrule
        
             \hlrow\textbf{UFC} &512$\times$ 512&Ori&\multicolumn{2}{c|}{Integrative Transformer}&\textbf{8.40}&\textbf{34.1} &\textbf{48.5}&\textbf{64.4}&\textbf{72.1} &\textbf{88.0} &\textbf{94.8} &\textbf{97.9} &\textbf{58.6} &\underline{81.2} &\textbf{50.4}&74.2\\

            \hlinewd{0.8pt}
    \end{tabular}
    }\vspace{-5pt}
    \caption{\textbf{Semantic matching results.}  
    }\label{tab:main_table}
    \end{center}\vspace{-15pt}
\end{table*}

\begin{table*}[]
    \centering
     
    \scalebox{0.51}{
    \begin{tabular}{l|cc|ccccc|c|c|ccccccc|c}
    \toprule
    \multirow{3}{*}{Methods}&\multirow{3}{*}{Feat.Agg.}&\multirow{3}{*}{Cost.Agg.} &\multicolumn{7}{c|}{HPatches Original} &\multicolumn{8}{c}{ETH3D}\\\cline{4-18}
    &&&\multicolumn{6}{c|}{AEPE $\downarrow$}&{PCK $\uparrow$}&\multicolumn{8}{c}{AEPE $\downarrow$}\\\cline{4-9}\cline{10-18}
    &&&I&II&III&IV&V&Avg.&5px& rate=3 & \multicolumn{1}{r}{rate=5} & rate=7 & rate=9 & rate=11 & rate=13 & rate=15&Avg.\\\midrule\midrule
    COTR &Trans. & -&-&-&-&-&-&{7.75}&{91.10} &1.66 & 1.82  & 1.97 & 2.13 & 2.27 & 2.41 & 2.61 & 2.12\\
		
   PUMP &- &4D Conv. &-&-&-&-&-&2.87&\underline{97.14}&1.77&2.81&2.39&2.39&3.56&3.87&4.57&3.05
			\\
     ECO-TR &Trans. &- &-&-&-&-&-&\underline{2.52}&90.85 &\underline{1.48}&\underline{1.61}&\underline{1.72}&\underline{1.81}&\underline{1.89}&\underline{1.97}&\underline{2.06} &\underline{1.87}
			\\
   COTR+Interp.  &Trans. & -&-&-&-&-&-&{7.98}&{86.33} &1.71&1.92&2.16&2.47&2.85&3.23&3.76&2.59
		\\
  \hlrow\textbf{UFC} + (C)&\multicolumn{2}{c|}{{Integrative Transformer}}&\textbf{0.87}&\textbf{1.29}&\textbf{1.37}&\textbf{3.19}&\textbf{1.92}&\textbf{1.73}&\textbf{98.76}&\textbf{1.45}&\textbf{1.59}&\textbf{1.64}&\textbf{1.76}&\textbf{1.82}&\textbf{1.90}&\textbf{1.95}&\textbf{1.73}\\

 \midrule
			GLU-Net & -&2D Conv.&1.55&12.66&27.54&32.04&52.47&25.05&{78.54} & 1.98          & 2.54                       & 3.49          & 4.24          & 5.61          & 7.55          & 10.78 & 5.17       \\
			GLU-Net-GOCor &-&Hand-crafted &\underline{1.29}&10.07&23.86&27.17&38.41&20.16&81.43 & 1.93          & 2.28                       & 2.64         & 3.01          & 3.62          & 4.79          & 7.80 & 3.72       \\

			DMP &2D Conv. &2D Conv. &3.21&15.54&32.54&38.62&63.43&30.64&63.21 & 2.43          & 3.31                     & 4.41          &5.56          & 6.93          & 9.55          & 14.20 & 6.62        \\ 
   PDCNet (MS) &-&2D Conv. &\textbf{1.15}&\underline{7.43}&\underline{11.64}&\underline{25.00}&\underline{30.49}&\underline{15.14}&\textbf{91.41}&{1.60}&{1.79}&{2.00}&{2.26}&{2.57}&{2.90}&{3.56}&{2.38}\\

   GMFlow&Trans. &- &4.72&26.46&40.75&62.49&79.80&42.85&69.50&1.64&1.86&2.12&\underline{2.36}&3.49&5.62&10.64&3.96\\
   COTR${\dagger}$ &Trans. &- &19.65&33.81&45.81&62.03&66.28&{45.52}&{5.10} &    8.76&9.86&11.23&12.44&13.77&14.94&16.09&12.44  \\
   PDC-Net+ (MS) &-&2D Conv. &-&-&-&-&-&-&-&\underline{1.58}&\underline{1.76}&\textbf{1.96}&\textbf{2.16}&\textbf{2.49}&\underline{2.73}&\underline{3.24}&\underline{2.27}\\

  \midrule
				
  		\hlrow\textbf{UFC} &\multicolumn{2}{c|}{{ Integrative Transformer}}&1.91&\textbf{6.13}&\textbf{5.62}&\textbf{6.36}&\textbf{19.44}&\textbf{7.88}&\underline{89.36}&\textbf{1.54}&\textbf{1.72}&\underline{1.99}&\underline{2.18}&\underline{2.58}&\textbf{2.66}&\textbf{3.01}&\textbf{2.24}
    
    \\\bottomrule
    \end{tabular}}\vspace{-5pt}\caption{\textbf{Geometric matching results.}  A higher scene label or rate, \textit{i.e.,} V or 15, consists of more difficult images with extreme geometric deformations. $\dagger$ : Dense evaluation without zoom-in technique and confidence thresholding. \textit{(C) : Confidence thresholding.} }\vspace{-10pt}
    \label{tab:hpatches}
\end{table*}

\subsection{Semantic Matching}
We first evaluate ours on semantic matching, where large intra-class variations and background clutters pose additional challenges to matching.  Dense zoom-in is not applied for semantic matching due to the low resolution of evaluation images. We use three standard benchmarks: SPair-71k~\citep{min2019spair}, PF-PASCAL~\citep{ham2017proposal} and PF-WILLOW~\citep{ham2016proposal}. We follow the evaluation protocol of~\citep{cho2022cats}. The results are summarized in Table~\ref{tab:main_table}, where UFC outperforms others for all the benchmarks at almost all PCKs, showing robustness to the above challenges. While VAT~\citep{hong2022cost} and NeMF~\citep{hong2022neural} perform better at $\alpha_\mathrm{bbox}$, we stress that VAT evaluates at higher resolution and NeMF specializes in fine-grained correspondences. Also, we note that our performance is slightly inferior to SCOT~\citep{liu2020semantic} and PWarPC~\citep{truong2022probabilistic} for $\alpha_{\text{bbox-kp}} = 0.1$, but this is compensated for by the superior performance at lower alpha, \textit{i.e.,} 0.05, and the fact that PF-PASCAL~\citep{ham2017proposal} and PF-WILLOW~\citep{ham2016proposal} are small-scale datasets with a limited number of image pairs. Moreover, we highlight that for SPair-71k~\citep{min2019spair}, the largest dataset in semantic correspondence with extreme viewpoint and scale difference, UFC outperforms competitors at all PCKs. %

\subsection{Geometric Matching}\label{5.3}
We next show that our method also performs very well in geometric matching. Following the evaluation protocol of~\citep{truong2021learning}, we report the results on HPatches~\citep{balntas2017hpatches} and ETH3D~\citep{schops2017multi} in Table~\ref{tab:hpatches}. From the results, our method clearly outperforms existing dense matching networks, including those that perform additional optimization~\citep{truong2020gocor,Hong_2021_ICCV,truong2021learning} and  inference strategies~\citep{jiang2021cotr,truong2021learning}, and a representative optical flow method, GMFlow~\citep{xu2021gmflow}. Note that UFC excels at finding correspondences under extreme geometric deformations, \textit{i.e.,} scene IV and V. Interestingly, as UFC outperforms others at all intervals of ETH3D that consist of image sequences with varying magnitudes of geometric transformations, this indicates that it can also perform well in optical flow settings. This is supported by the results of GMFlow~\citep{xu2021gmflow}, where we consistently achieve better performance. To ensure a fair comparison to COTR~\citep{jiang2021cotr} and its follow-up works~\citep{revaud2022pump,tan2022eco}, we present results from a variant of our method, denoted as (C). Moreover, we also include COTR$\dagger$ to represent a truly dense version of COTR~\citep{jiang2021cotr}, and we observe that UFC clearly performs better.

\subsection{Quantitative comparison between aggregation strategies}\label{quan}
\setlength\intextsep{0pt}
\begin{wraptable}{r}{7.0cm}

    \centering
    \scalebox{0.6}{
   \begin{tabular}{ll|cc}
        \hlinewd{0.8pt}
        &\multirow{2}{*}{} &HPatches &SPair-71k \\
        &&AEPE $\downarrow$&$\alpha_{\text{bbox}}$ = 0.1 $\uparrow$ \\
        \midrule
        \textbf{(I)} &Feature self-att.  & 78.2&36.1\\
         \textbf{(II)} &  Feature self-att. and cross-att. &59.9&38.5 \\
        \textbf{(III)} & Cost self-att.  &51.8&33.6\\\midrule
        \textbf{(IV)} &  Feature self-att. + cost self-att. &36.8 &51.7 \\
        \textbf{(V)} &   Feature self- and cross-att. + cost self-att. &26.2&56.5\\
        \hlinewd{0.8pt}
\end{tabular}}%
\vspace{-5pt}
        \caption{\textbf{Comparison of aggregation strategies.}}
    \label{tab:aggr}
   
\end{wraptable}
Figures~\ref{PCA} and~\ref{fig:multi-cost} qualitatively show that the information both aggregations exploit and the output they generate differ. Here, we empirically show that these lead to appreciable performance differences. Table~\ref{tab:aggr} (I-III) compares different aggregation strategies that are trained and evaluated on both tasks. Note that for these variants, we do not include any module for boosting performance, e.g., coarse-to-fine, multi-level or
multi-scale features, and maintain a similar number of learnable parameters. Pytorch-like pseudocodes and additional visualizations are given in the supplementary material. From the results, we find that each aggregation strategy yields apparently different results in two tasks as reported in (I-III). A particularly illustrative comparison is (I) vs.~(III), where only self-attention is performed on features or the cost volume, which clearly differentiate them. As expected, we also observe that performing cross-attention improves performance.    

In the last two rows, we report the results of variants that utilize both feature and cost aggregation. (V) is our integrative aggregation, and (IV) is a na\"ive sequential aggregation where feature aggregation is followed by cost aggregation.
They both achieve large performance boosts, as expected since the resultant cost volume in Fig.~\ref{fig:multi-cost} (f) and (h) includes less noisy and more concentrated scores compared to (c-e), while integrative aggregation clearly performs better. From these quantitative comparisons, we highlight that the two types of aggregation serve different purposes that lead to apparent performance differences, and the potential benefits arising from their relationship can be further exploited with our proposed design.

\subsection{Ablation Study}\label{sec:4.5}

\setlength\intextsep{0pt}
\begin{wraptable}{r}{7.0cm}
    \centering
    \scalebox{0.63}{
    \begin{tabular}{ll|cc}
        \hlinewd{0.8pt}
        &\multirow{2}{*}{Components} &HPatches &SPair-71k \\
        &&AEPE $\downarrow$&$\alpha_{\text{bbox}}$ = 0.1 $\uparrow$ \\
        \midrule
        
        \textbf{(I)} &   Baseline &36.8&51.7\\
        \textbf{(II)} &  Integrative self-att. &32.8 &54.7\\
        \textbf{(III)}& Integrative self- and cross-att. &22.8&58.4\\
        \textbf{(IV)}&  
        + matching distribution. &{18.7}&59.9\\
        \textbf{(V)}&+ hierarchical processing &10.9&64.4\\
    \textbf{(VI)}&  + dense zoom-in & 7.88&-\\
        \hlinewd{0.8pt}
\end{tabular}}%
\vspace{-5pt}
        \caption{\textbf{Component ablation study.} }\label{tab:aggregate}

\end{wraptable}
In this ablation study, we verify the need for each component of our method. Table~\ref{tab:aggregate} presents quantitative results of each variant on both geometric and semantic matching. The baseline represents a variant equipped with self- and cross-attention on feature maps. From (II) to (VI), each proposed component is progressively included. We control the number of learnable parameters for each variant to be similar.

Comparing (I) and (II), we find that the proposed integrative self-attention layer benefits from joint aggregation of features and cost volume, achieving improved performance on both tasks. We next find that each component clearly helps to boost performance. Interestingly, we find dramatic improvements when cross-attention is included (III), indicating that explicit conditioning between input images is helpful. This further enhanced by using the aggregated cost volume as a cross-attention map (IV).

\section{Conclusion}

In this paper, we introduced a simple yet effective dense matching approach, Unified Feature and Cost Aggregation with Transformers (UFC), that capitalizes on the distinct advantages of the two types of aggregation. We further devise an enhanced aggregation through cross-attention with matching distribution. This method is formulated in a coarse-to-fine manner, yielding an appreciable performance boost. We have shown that our approach exhibits high speed and efficiency and that it surpasses all other existing works on several benchmarks, establishing new state-of-the-art performance. %

\noindent\textbf{Acknowledgement} This research was supported by the MSIT, Korea (IITP-2024-2020-0-01819, ICT Creative Consilience Program, RS-2023-00227592, Development of 3D Object Identification Technology Robust to Viewpoint Changes), and National Research Foundation of Korea (NRF-2021R1A6A1A03045425).

\clearpage

% \title{Unifying Feature and Cost Aggregation with Transformers for Semantic and Visual Correspondence }

% \author{Sunghwan Hong\thanks{equal contribution} \   , Seokju Cho$^*$, Seungryong Kim$^{\dagger}$\thanks{$^\dagger$Corresponding author} \\
% Korea University\\
% \texttt{\{sung$\_$hwan,seokju$\_$cho,seungryong$\_$kim\}@korea.ac.kr} \\
% \And
% Stephen Lin \\
% Microsoft Research Asia\\
% \texttt{stevelin@microsoft.com} \\
% }

% \maketitle
\appendix

\section*{Appendix}

In the following, we first provide more implementation details in Section~\ref{A}. Then, we provide details on evaluation metrics and datasets in Section~\ref{B}. We then explain the training procedure in more depth in Section~\ref{C}. Subsequently, we provide additional experimental results and ablation study in Section~\ref{D}. We then provide clarifications to the evaluation procedure adopted by COTR~\citep{jiang2021cotr} and its follow-up works in Section~\ref{E}. Finally, we present qualitative results for all the benchmarks in Section~\ref{G} and a discussion of future work in Section~\ref{H}.

\section{Implementation Details}\label{A}
\subsection{Network Architectures}
To extract features, we use ResNet-101~\citep{he2016deep} for semantic matching, and VGG-16~\citep{simonyan2014very} for geometric matching, consistent with prior works~\citep{min2020learning,min2021convolutional,cho2021semantic,hong2022cost,truong2020glu,truong2020gocor,truong2021learning}. We select three feature maps from the last convolutional block, namely $\mathrm{Conv}$3$\_$x, $\mathrm{Conv}$4$\_$x, and $\mathrm{Conv}$5$\_$x, with channel dimensions of 2048, 1024, and 512, respectively. To reduce computational complexity, we project each feature map to smaller dimensions of 384, 256, and 128, respectively, before constructing a cost volume and passing it to our integrative aggregation block. Bilinear interpolation is used to adjust the  spatial dimensions of intermediate outputs. The resolutions at each levels $l = 1,2,3$ are 16$\times$ 16, 32$\times$ 32 and 64$\times$ 64, respectively. For the final output flow map, we use a soft-argmax operator with temperature set to 0.02. 

\subsection{Other Implementation Details}
\paragraph{COTR Implementation Details.}
In the main table, we report the results of COTR~\citep{jiang2021cotr} without zoom-in and confidence thresholding. Here, we provide the implementation details for how we obtained the results. 

To adapt the input pair of images for use in our evaluation, we resize them to 256$\times$ 256, which matches the resolution used by COTR~\citep{jiang2021cotr}. Rather than selecting sparse coordinates for finding correspondences, we input all coordinates defined at the original resolution of the images, resulting in dense correspondences. This approach involves feed-forwarding all coordinates in parallel, which speeds up the process. We then compute the average endpoint error (AEPE) for all correspondences, masking any invalid correspondences as per conventional evaluation protocols~\citep{truong2020glu,truong2020gocor,truong2021learning}. Note that our evaluation does not take into account the zoom-in technique used in COTR. We also evaluate dense correspondence rather than the original sparse or quasi-dense evaluation method adopted by COTR~\citep{jiang2021cotr}, which we detail in Section~\ref{E}.

\begin{figure*}
    \centering
    \includegraphics[width=1\linewidth]{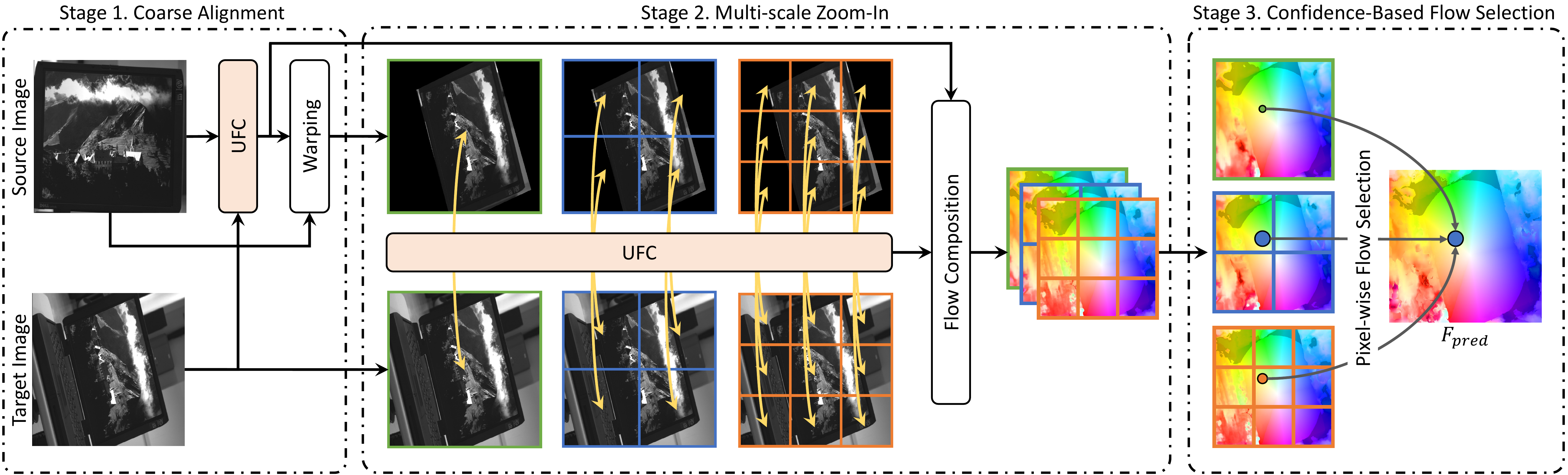}\hfill\\
    \caption{\textbf{Pipeline of dense zoom-in technique.}}
    \label{zoom}\vspace{-10pt}
\end{figure*}

\paragraph{GMFlow Inference Details.}
To evaluate GMFlow on HPatches~\citep{balntas2017hpatches} and ETH3D~\citep{zhao2021multi}, we use the weights given at the official implementation page. However, processing HPatches at its original resolution requires an enormous amount of memory, even when using a high-end GPU such as the 80GB A100. Therefore, we interpolate the input image pairs using bilinear interpolation to match the size of the crops used during training (320$\times$ 896), before feeding them into GMFlow~\citep{xu2021gmflow}. We use the weights trained on Sintel~\citep{butler2012naturalistic} with refinement strategy to obtain the best results.

Since there are numerous hyperparameters that can affect model performance, \textit{i.e.,} padding factor, window sizes for Swin Transformer~\citep{liu2021swin} and scale factor, we choose the same hyperparameters $u$ that were used to obtain the pre-trained weight. Specifically, we set padding factor to 32, upsample factor to 4, scale factor to 2, attention split list to (2,8), correlation radius list to (-1,4) and flow propagation list to (-1,1). We keep all other hyperparameters at their default values. We then evaluate the model using the same evaluation procedure as in previous works~\citep{truong2020glu,truong2020gocor,truong2021learning}. \vspace{-10pt}

\paragraph{PDC-Net Inference Details.}
To evaluate PDC-Net~\citep{truong2021learning} on HPatches~\citep{balntas2017hpatches} and ETH3D~\citep{schops2017multi}, we simply use the pre-trained weights and the official implementation codes. Note that the only hyperparameters we change at inference are GOCor~\citep{truong2020gocor} hyperparameters, for which we set the number of iterations for global and local correlation map optimization to 3 and 7, respectively. The rest of the hyperparameters remain as the default values.

\subsection{Dense Zoom-In}
We provide an overview of dense zoom-in used at inference phase, in Fig.~\ref{zoom}.
\section{Evaluation Metrics and Datasets}\label{B}

\subsection{Geometric Matching}
\paragraph{Compared Methods.}
There are two groups of methods we compare to. The first group includes COTR~\citep{jiang2021cotr}, PUMP~\citep{revaud2022pump} and ECO-TR~\citep{tan2022eco}. The second group includes GLU-Net~\citep{truong2020glu}, GOCor~\citep{truong2020gocor}, DMP~\citep{Hong_2021_ICCV}, PDC-Net~\citep{truong2021learning}, PDC-Net+~\citep{truong2021pdc+}, and they are trained in a self-supervised manner on either DPED-CityScape-ADE~\citep{ignatov2017dslr,cordts2016cityscapes,zhou2019semantic} or MegaDepth~\citep{li2018megadepth} except for GMFlow~\citep{xu2021gmflow}. For the second group, the evaluation is done using \url{https://github.com/PruneTruong/DenseMatching}.
\vspace{-10pt}
\paragraph{Evaluation Metric.} For the evaluation metric, we use the average end-point error (AEPE), computed by averaging the Euclidean distance between the ground-truth and estimated flow, and percentage of correct keypoints (PCK), computed as the ratio of estimated keypoints within a threshold of the ground truth to the total number of keypoints. More specifically, AEPE is computed using the following equation: $\|F_\mathrm{GT}-F_\mathrm{pred}\|_{2}$, where $F_\mathrm{GT}$ represents a ground-truth dense flow map and $F_\mathrm{pred}$ represents a dense predicted flow map. \vspace{-10pt}

\paragraph{HPatches.} Hpatches~\citep{balntas2017hpatches}  consists of images with different views of the same scenes. Each sequence contains a source and five target images with different viewpoints and the corresponding ground-truth flows. Generally, the later scenes, \textit{i.e.,} IV and V, consist of more challenging target images. We use images of high resolutions ranging from 450 $\times$ 600 to 1,613 $\times$ 1,210. \vspace{-10pt}

\paragraph{ETH3D.}
Unlike Hpatches~\citep{balntas2017hpatches}, ETH3D~\citep{schops2017multi} consists of real 3D scenes, where the image transformations are not constrained to a homography. This multi-view dataset contains 10 image sequences ranging from 480 $\times$ 752 to 514 $\times$ 955. The authors of ETH3D~\citep{schops2017multi} additionally provide a set of sparse image correspondences, for which we follow the protocol of~\citep{truong2020glu} by sampling the image pairs at different intervals to evaluate on varying magnitudes of geometric transformations. We evaluate on 7 intervals in total, each interval containing approximately 500 image pairs, or 600K to 1000K correspondences. Generally, the image pairs are more challenging at a higher rate, \textit{i.e.,} rate 13 or 15, as shown in Fig.~\ref{eth1} and Fig.~\ref{eth2}.

\subsection{Semantic Matching}
\paragraph{Compared Methods.}
We compare our methods to semantic matching methods, which are all trained in a supervised manner using ground-truth keypoints. DHPF~\citep{min2020learning}, SCOT~\citep{liu2020semantic}, CHM~\citep{min2021convolutional}, CATs~\citep{cho2021semantic}, MMNet~\citep{zhao2021multi}, PWarpC-NC-Net~\citep{truong2022probabilistic,Rocco18b}, SCorrSAN~\citep{huang2022learning}, VAT~\citep{hong2022cost}, CATs++~\citep{cho2022cats}, TransforMatcher~\citep{kim2022transformatcher} and NeMF~\citep{hong2022neural} are trained with SPair-71k~\citep{min2019spair} when evaluated on SPair-71k and they are trained on PF-PASCAL~\citep{ham2017proposal} when evaluated on PF-PASCAL and PF-WILLOW~\citep{ham2016proposal}. Note that all the methods adopt ResNet-101 except for MMNet~\citep{zhao2021multi}. \vspace{-10pt}

\paragraph{Evaluation Metric.}
For semantic matching, following~\citep{Rocco18b,min2019hyperpixel,min2020learning,cho2021semantic,9933865},  we transfer the annotated keypoints in the source image to the target image using the dense correspondence between the two images. The percentage of correct keypoints (PCK) is computed for evaluation. Note that higher PCK values are better. Concretely, given predicted keypoint $k_\mathrm{pred}$ and ground-truth keypoint $k_\mathrm{GT}$, we count the number of predicted keypoints that satisfy the following condition: $d( k_\mathrm{pred},k_\mathrm{GT}) \leq \alpha \cdot \mathrm{max}(H,W)$, where $d(\,\cdot\,)$ denotes Euclidean distance; $\alpha$ denotes a threshold value; $H$ and $W$ denote height and width of the object bounding box or the entire image. Note that we additionally reported results of PCK @ $\alpha_{\text{bbox-kp}}$ to compensate for the fact that~\citep{min2020learning,min2021convolutional,cho2021semantic,hong2022cost,huang2022learning} chose thresholds different from other works~\citep{liu2020semantic,lee2019sfnet,Rocco18b,rocco2017convolutional} for PF-WILLOW~\citep{ham2016proposal}.\vspace{-10pt}
\paragraph{SPair-71k.}
SPair-71k~\citep{min2019spair} is a large-scale benchmark for semantic correspondence, which consists of 18 object categories of 70,958 image pairs with extreme and diverse viewpoints, scale variations, and rich annotations for each image pair. Ground-truth annotations for object bounding boxes, segmentation masks and keypoints are available. For the evaluation, we follow the conventional evaluation protocol~\citep{min2019hyperpixel} of using a test split of 12,234 image pairs. \vspace{-10pt}
\paragraph{PF-PASCAL and PF-WILLOW.}
PF-PASCAL~\citep{ham2017proposal} is a dataset introduced as an extension for PF-WILLOW~\citep{ham2016proposal}. It consists of 1,351 image pairs of 20 image categories, while PF-WILLOW~\citep{ham2016proposal} consists of 900 image pairs of 4 image categories. PF-PASCAL~\citep{ham2017proposal} is a more challenging dataset than others, \textit{i.e.,} TSS~\citep{taniai2016joint} or PF-WILLOW~\citep{ham2016proposal}, for semantic correspondence evaluation, as it additionally exhibits large appearance, scene layout, scale and clutter changes. For evaluation, we use the test split of PF-PASCAL~\citep{ham2017proposal} and PF-WILLOW~\citep{ham2016proposal}. 

\section{Training Details}\label{C}
In this section, we provide training details for both semantic and geometric matching. We employ an Intel Core i7-10700 CPU and RTX-3090 GPUs for training. 

\subsection{Dense Geometric Matching}
For geometric matching, we adopt two-stage training. At the first stage, we freeze the backbone network and only train UFC using DPED-CityScape-ADE~\citep{ignatov2017dslr,cordts2016cityscapes,zhou2019semantic}. This stage is similar to the training procedure of GOCor-GLU-Net~\citep{truong2020gocor}. More specifically, due to the limited amount of dense correspondence data, most matching networks resort to self-supervised training, where synthetic warps provide dense correspondences. To this end, we adopt the same training procedure to GOCor-GLU-Net~\citep{truong2020gocor} that consists of pairs of images created by synthetically warping the image according to random affine, homography or TPS transformations. We crop the images to 512$\times$ 512, and in total, we use 40K image pairs for the first stage of training. We set the learning rate to 3e$^{-4}$, use AdamW~\citep{loshchilov2017decoupled} and iterate for 50 epochs with the batch size set to 16. We freeze the backbone in this stage.

For the second stage, we continue from the best model from the first stage, which was chosen by cross-validation. For the dataset, we use the MegaDepth dataset, which consists of 196 different scenes reconstructed from about 1M internet images using COLMAP~\citep{schonberger2016structure} and combine this with the synthetic data. For training, we sample up to 500 random images from 150 different scenes in which the  overlap is at least 30\% with the sparse SfM point cloud. We also include random independently moving objects sampled from the COCO~\citep{lin2014microsoft} dataset on top of the synthetic data. Moreover, we also utilize perturbation data as we found it beneficial to include in the dataset. Finally, for the validation dataset, we sample up to 80 random images pairs from 25 different scenes. We resize the images to 512$\times$ 512 in consistency with the first stage. For the second stage, we train the whole network, set the learning rate to 1e$^{-4}$, use AdamW~\citep{loshchilov2017decoupled} and iterate for 175 epochs with the batch size set to 16. 

\paragraph{Dense Semantic Matching}
To ensure a fair comparison, following~\citep{min2021convolutional,cho2021semantic}, when evaluating on SPair-71k~\citep{min2019spair} we train the proposed method on the training split of SPair-71k~\citep{min2019spair}, and when evaluating on PF-PASCAL~\citep{ham2017proposal} and PF-WILLOW~\citep{ham2016proposal} we train on the training split of PF-PASCAL~\citep{ham2017proposal}. We only train the UFC module and freeze the backbone network. We apply random augmentation~\citep{buslaev2020albumentations} as done in~\citep{cho2021semantic}. We set the learning rate to 1e$^{-4}$, use AdamW~\citep{loshchilov2017decoupled} as an optimizer, set the batch size to 24, and iterate for 50 epochs for SPair-71k~\citep{min2019spair} and 300 epochs for PF-PASCAL~\citep{ham2017proposal}. The best model is obtained through cross-validation.

\section{Additional Quantitative Results and Ablation Study}\label{D}

\begin{figure}
  \centering
    \includegraphics[width=0.45\linewidth]{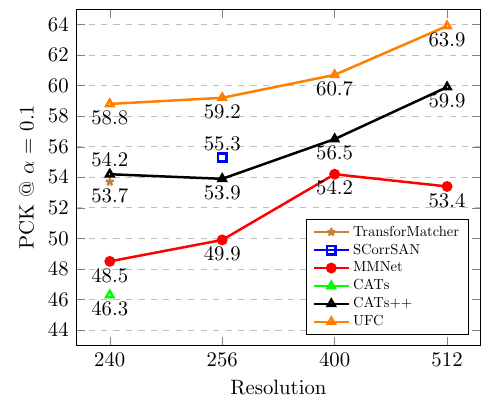}\hfill\\
    \vspace{-10pt}\caption{\textbf{Image resolution ablation.}}
    \label{reso}
\vspace{-10pt}
\end{figure}

\paragraph{Ablation study on image resolution.}
Here, we show additional results of our method UFC trained and evaluated at different resolutions on SPair-71k~\citep{min2019spair}. As done in CATs++~\citep{9933865}, we train and evaluate at 240, 256, 400 and 512 to directly compare with competitors, each of which train and evaluate at different resolutions,~\textit{i.e.,} 240 for CHM~\citep{min2021convolutional}, 256 for CATs~\citep{cho2021semantic}, 400 for PMNC~\citep{lee2021patchmatch} and 512 for CATs++~\citep{9933865}.   The results are shown in Fig.~\ref{reso}, where UFC still outperforms the competitors. \vspace{-10pt}

\paragraph{Effects of varying $k$.}
In Table~\ref{k}, we summarize the effects of varying $k$, a hyperparameter used for dense zoom-in at inference. HPatches~\citep{balntas2017hpatches} is used in measuring memory and run time. We take the maximum GPU memory utilization and average the run time. From the experiments, we find that varying $k$ has minor effects on performance while it has a large influence on memory and run time.  This means that input images having resolutions similar to HPatches~\citep{balntas2017hpatches} and ETH3D~\citep{schops2017multi} do not require higher $k$ that leads to unnecessarily increased memory and run time; rather, smaller $k$ should be chosen. Note that to measure the maximum GPU utilization at $k = (4,5,6)$, we use 80GB A100 as RTX-3090 lacks GPU capacity for this configuration. Because of this, we omit the run-time, as it is unfair to compare with other configurations. For semantic matching, we empirically find that varying $k$ barely has an impact on the performance, while increasing the complexity. This is likely due to the relatively small resolutions of the image pairs in the standard benchmarks~\citep{min2019spair,ham2016proposal,ham2017proposal}.

\section{Discussions}\label{E}
\paragraph{Sparse and Quasi-Dense Evaluation.}
In this section, we clarify the difference between the evaluation procedures adopted by COTR~\citep{jiang2021cotr} and its follow-up works~\citep{revaud2022pump,tan2022eco} to those of other existing works~\citep{Hong_2021_ICCV,truong2020glu,truong2020gocor,truong2021learning,melekhov2019dgc,shen2020ransac}. COTR~\citep{jiang2021cotr} is one of the first works to use Transformer to find correspondences between images. Its follow-up works, including ECO-TR~\citep{tan2022eco} and PUMP~\citep{revaud2022pump}, extend the work by improving performance or speed, or by reducing computations. These works attained state-of-the-art performance that significantly surpass the previous works, highlighting the effectiveness of these methods.

However, in order to compare with existing dense matching networks on dense correspondence datasets~\citep{balntas2017hpatches,schops2017multi}, they first find sparse correspondences based on the confidence scores, where the correspondences below certain thresholds are discarded as mentioned in COTR~\citep{jiang2021cotr}. Using only the sparse correspondences, AEPE or PCK is computed and compared to other works, which means that most of the erroneous correspondences that can significantly affect the metrics are not taken into account. On top of this, the densification is performed using only the confident sparse correspondences, and this version is indicated as ``+interp" or dense version in their papers. The metrics are then calculated using only the points within the convex hull, and the points outside of the interpolated regions are discarded in the evaluation. 
We find that this differs from conventional dense evaluation. Instead, this is more to be classified as ``semi-dense" or ``quasi-dense", which should be evaluated separately from existing dense methods.
% Although PUMP~\citep{revaud2022pump} presents an extrapolation method to produce fully dense correspondence in their official implementation, the results reported in the paper appear to be computed using the confident quasi-dense flow maps, which clearly give it an apparent advantage similar to COTR~\citep{jiang2021cotr} and ECO-TR~\citep{tan2022eco}. Qualitative examples are shown in Fig.~\ref{cotr1}. We find that in the evaluation, COTR and its follow-up works have advantages over other existing dense matching works.
\begin{table}[]
    \centering
    \scalebox{0.7}{
   \begin{tabular}{c|cc|cc}
        \hlinewd{0.8pt}
        
     \multirow{2}{*}{k}  &\multicolumn{2}{c|}{AEPE}&Memory & Run-time \\\cline{2-5}
        &HPatches~\citep{balntas2017hpatches}&ETH3D~\citep{schops2017multi}&[GB] & [s] \\
        \midrule
     (2, 3) & 7.79 & 2.57&{4.9} &{3.13}  \\
     (2, 3, 4) & 7.90& 2.49& 6.2& 6.05 \\
     (3, 4) & 7.89& 2.47& 6.0&5.32 \\
     (3, 4, 5) &7.88 &2.33&14.8&4.63\\
     (4, 5) & 7.99& 2.42 & 14.2& 8.13\\
     (4, 5, 6) &7.90 &2.33  &32.5 &- \\

        \hlinewd{0.8pt}
\end{tabular}}%
\vspace{5pt}
    \caption{\textbf{Partition ablation.}}
    \label{k}
   
\end{table}
% Although COTR~\citep{jiang2021cotr} explicitly mentions in a footnote that their performance cannot directly be compared to existing dense methods, the presentation in the paper may lead to some confusion that complicates comparisons, especially when PUMP~\citep{revaud2022pump} and ECO-TR~\citep{tan2022eco} do not address this. Therefore, we provide these clarifications to reduce further potential confusion.

\section{Qualitative Results}\label{G}
\begin{figure*}
    \centering
    \includegraphics[width=1.0\linewidth]{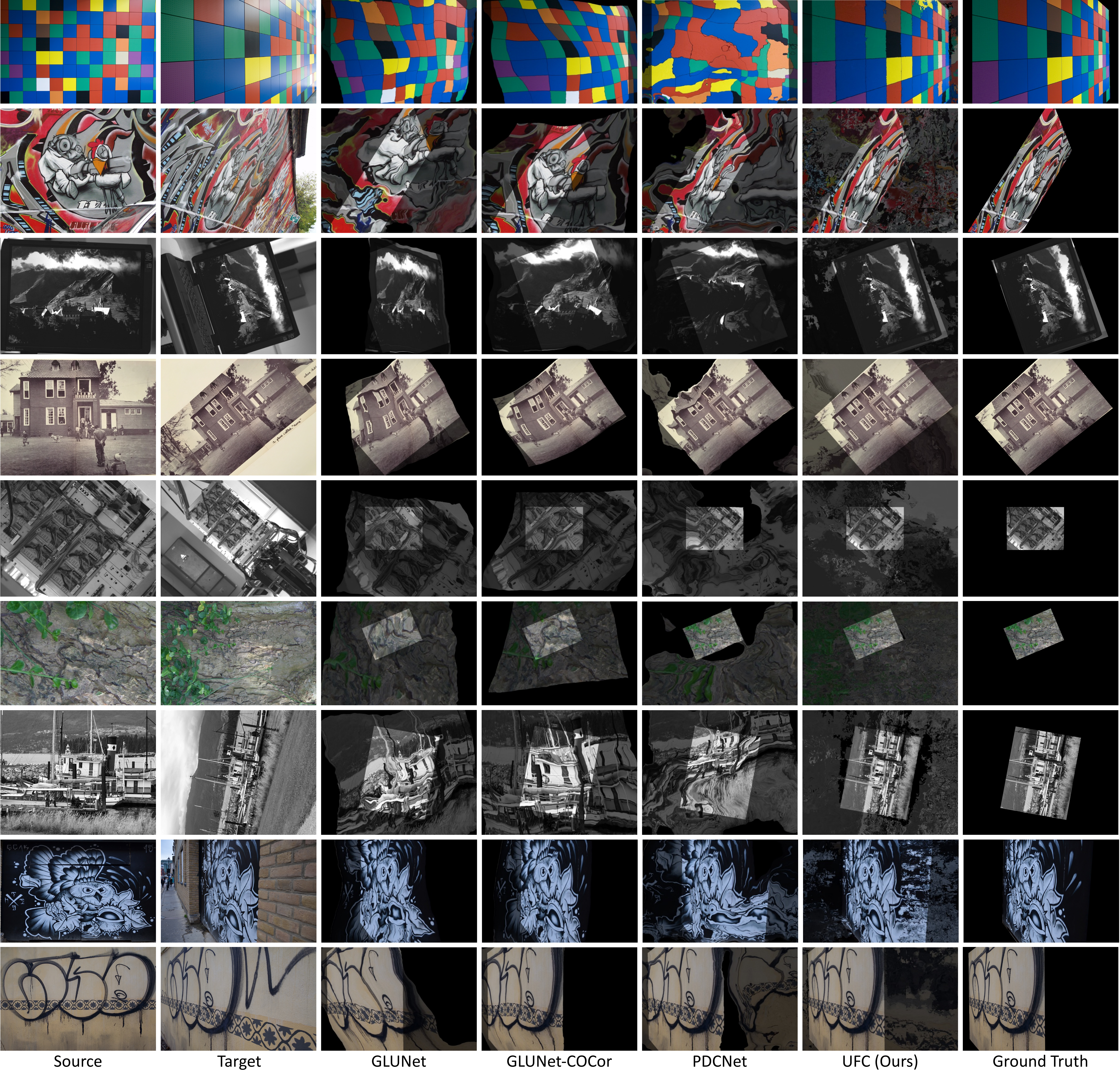}\hfill\\
    \caption{\textbf{Qualitative results on HPatches~\citep{balntas2017hpatches}.}}
    \label{hp}\vspace{-10pt}
\end{figure*}

\begin{figure*}
    \centering
    \includegraphics[width=1.0\linewidth]{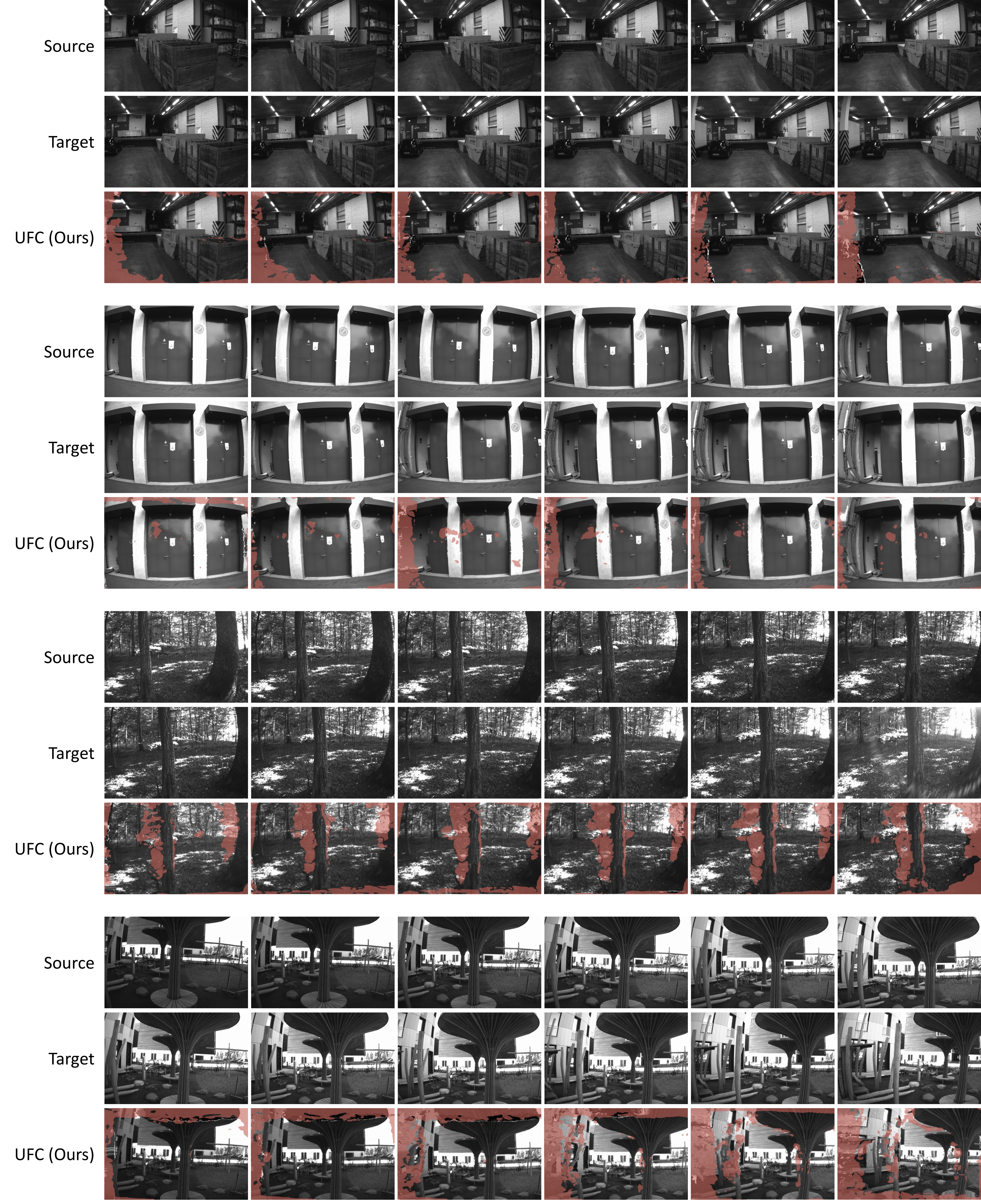}\hfill\\
    \caption{\textbf{Qualitative results on ETH3D~\citep{schops2017multi}.}}
    \label{eth1}\vspace{-10pt}
\end{figure*}

\begin{figure*}
    \centering
    \includegraphics[width=1.0\linewidth]{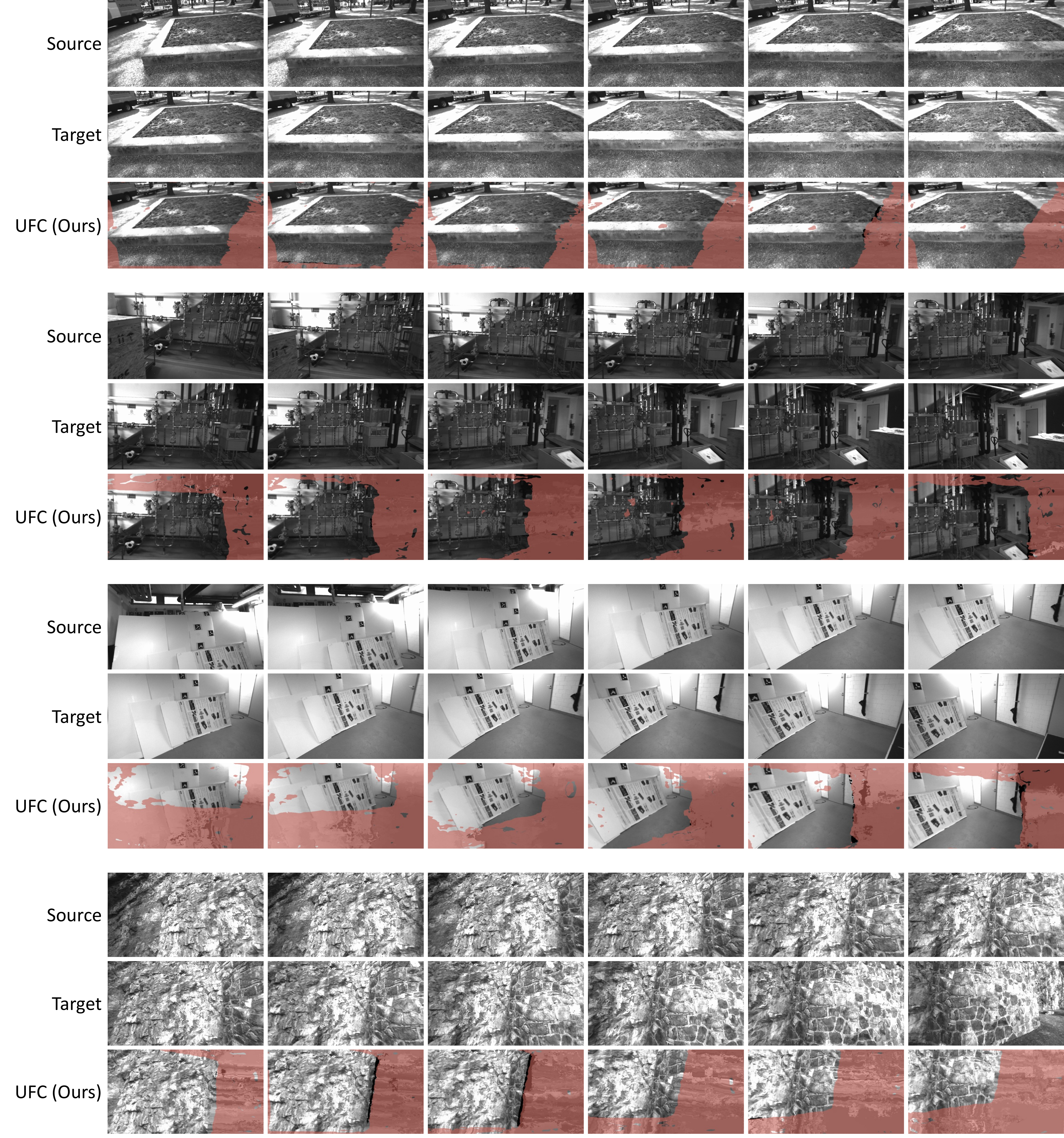}\hfill\\
    \caption{\textbf{Qualitative results on ETH3D~\citep{schops2017multi}.} }
    \label{eth2}\vspace{-10pt}
\end{figure*}

\begin{figure*}
    \centering
    \includegraphics[width=0.95\linewidth]{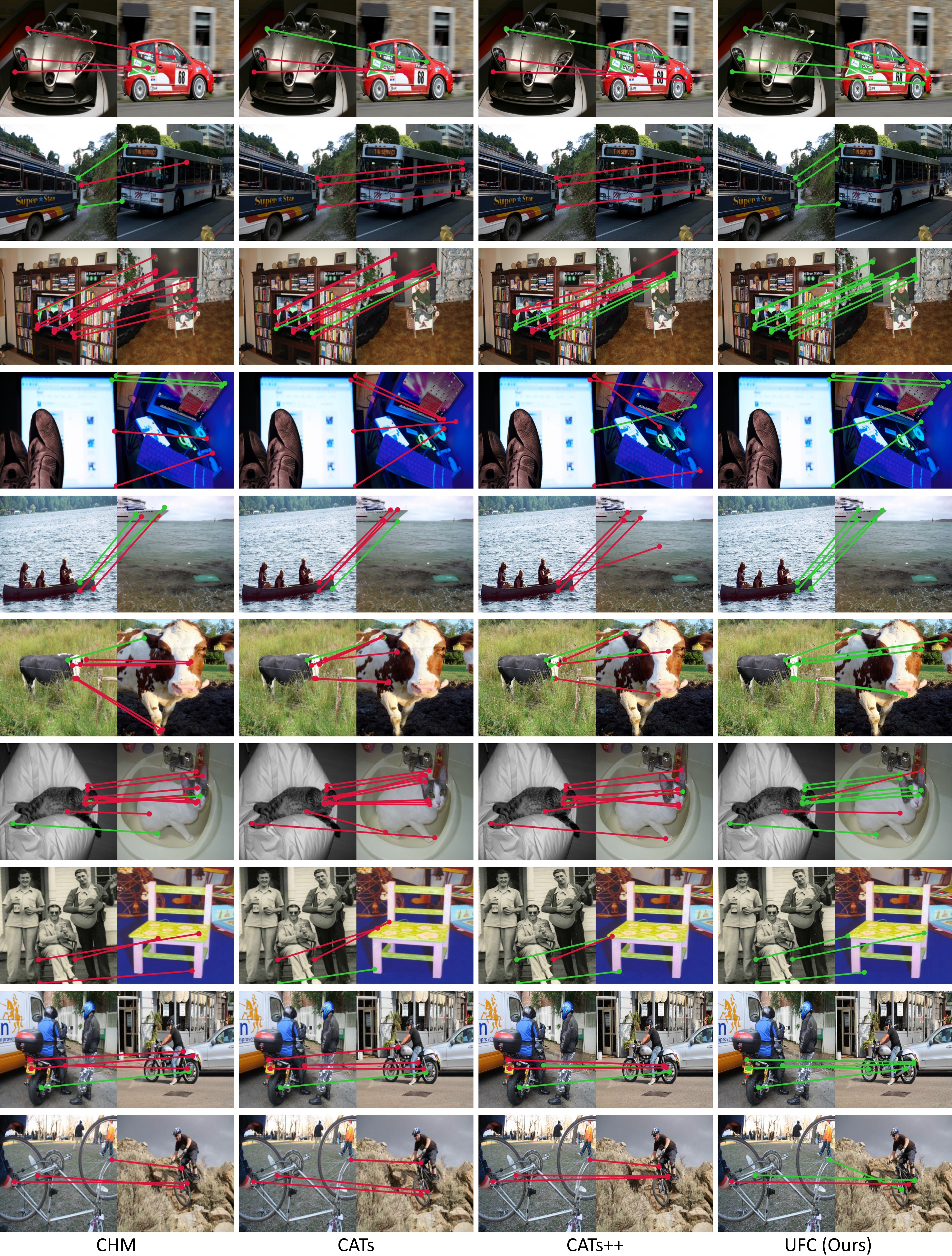}\hfill\\
    \caption{\textbf{Qualitative results on SPair-71k~\citep{min2019spair}:} keypoints transfer results by CHM~\citep{min2021convolutional}, CATs~\citep{cho2021semantic}, CATs++~\citep{9933865} and ours. Note that green and red lines denote correct and wrong predictions, respectively, with respect to the ground-truth.}
    \label{spair1}\vspace{-10pt}
\end{figure*}

\begin{figure*}
    \centering
    \includegraphics[width=0.95\linewidth]{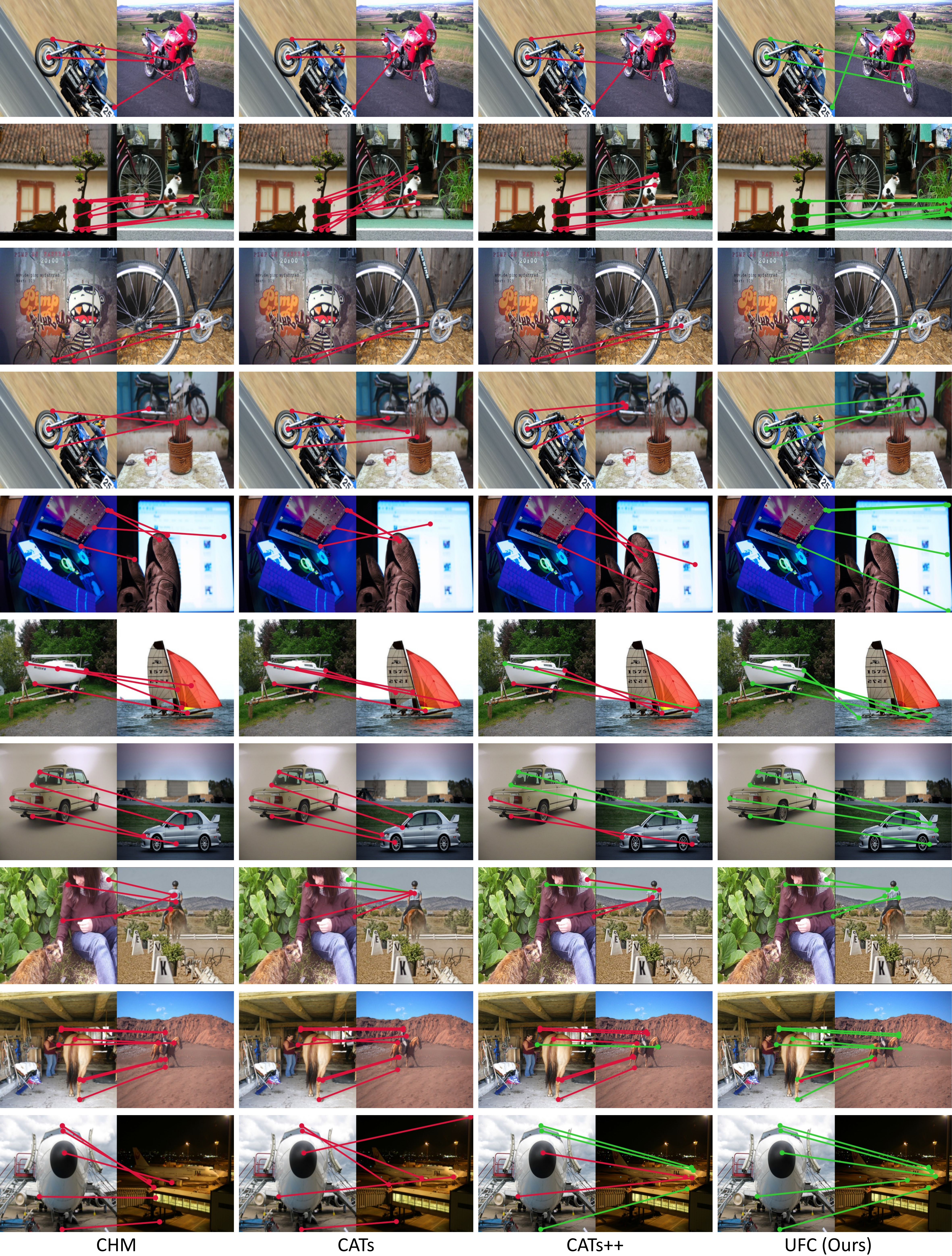}\hfill\\
    \caption{\textbf{Qualitative results on SPair-71k~\citep{min2019spair}:} keypoints transfer results by CHM~\citep{min2021convolutional}, CATs~\citep{cho2021semantic}, CATs++~\citep{9933865} and ours.}
    \label{spair2}\vspace{-10pt}
\end{figure*}

\begin{figure*}
    \centering
    \includegraphics[width=0.8\linewidth]{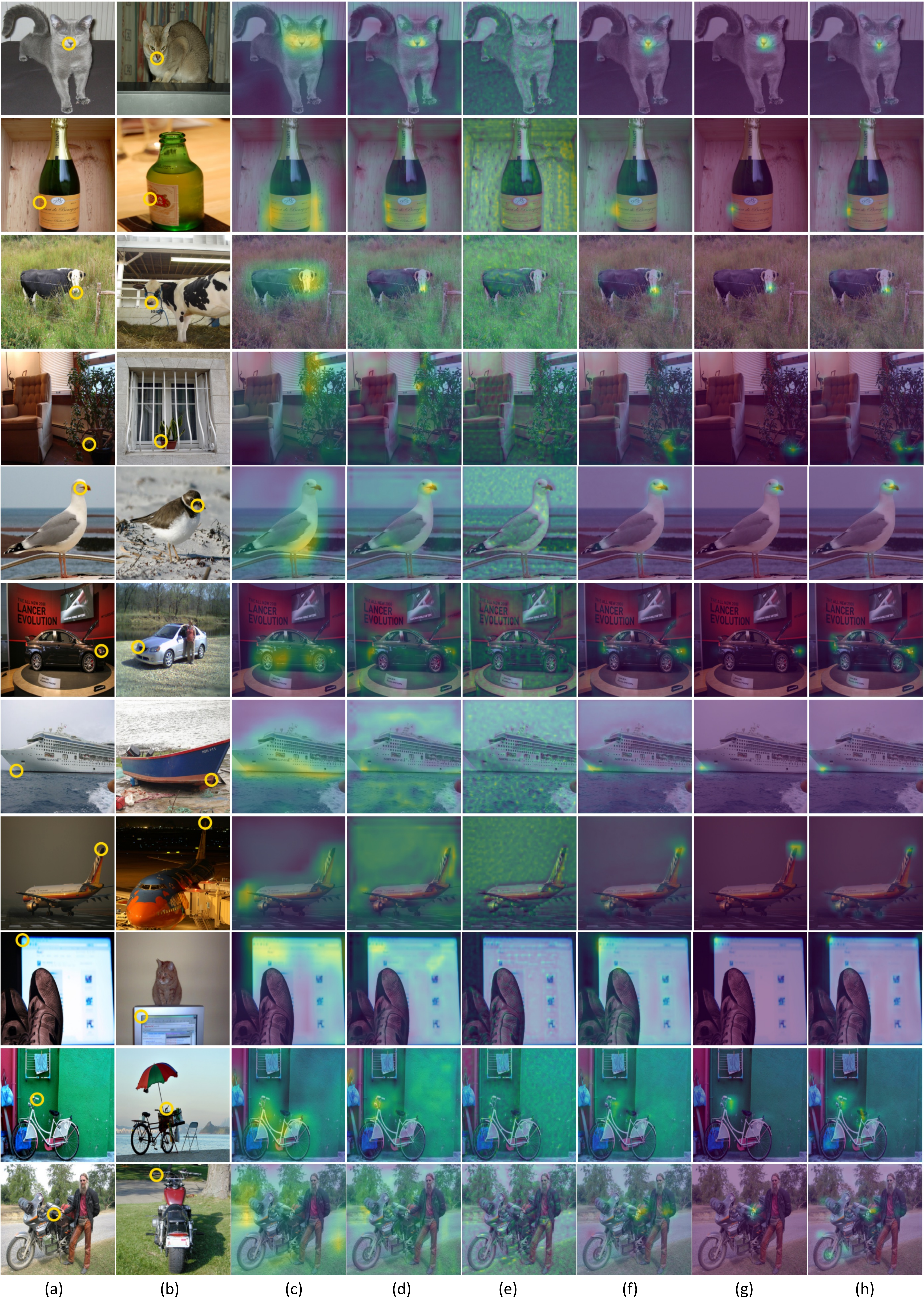}\hfill\\
    \caption{\textbf{Visualization of attention maps.} }
    \label{attention_supp}\vspace{-10pt}
\end{figure*}

\begin{figure*}
    \centering
    \includegraphics[width=1.0\linewidth]{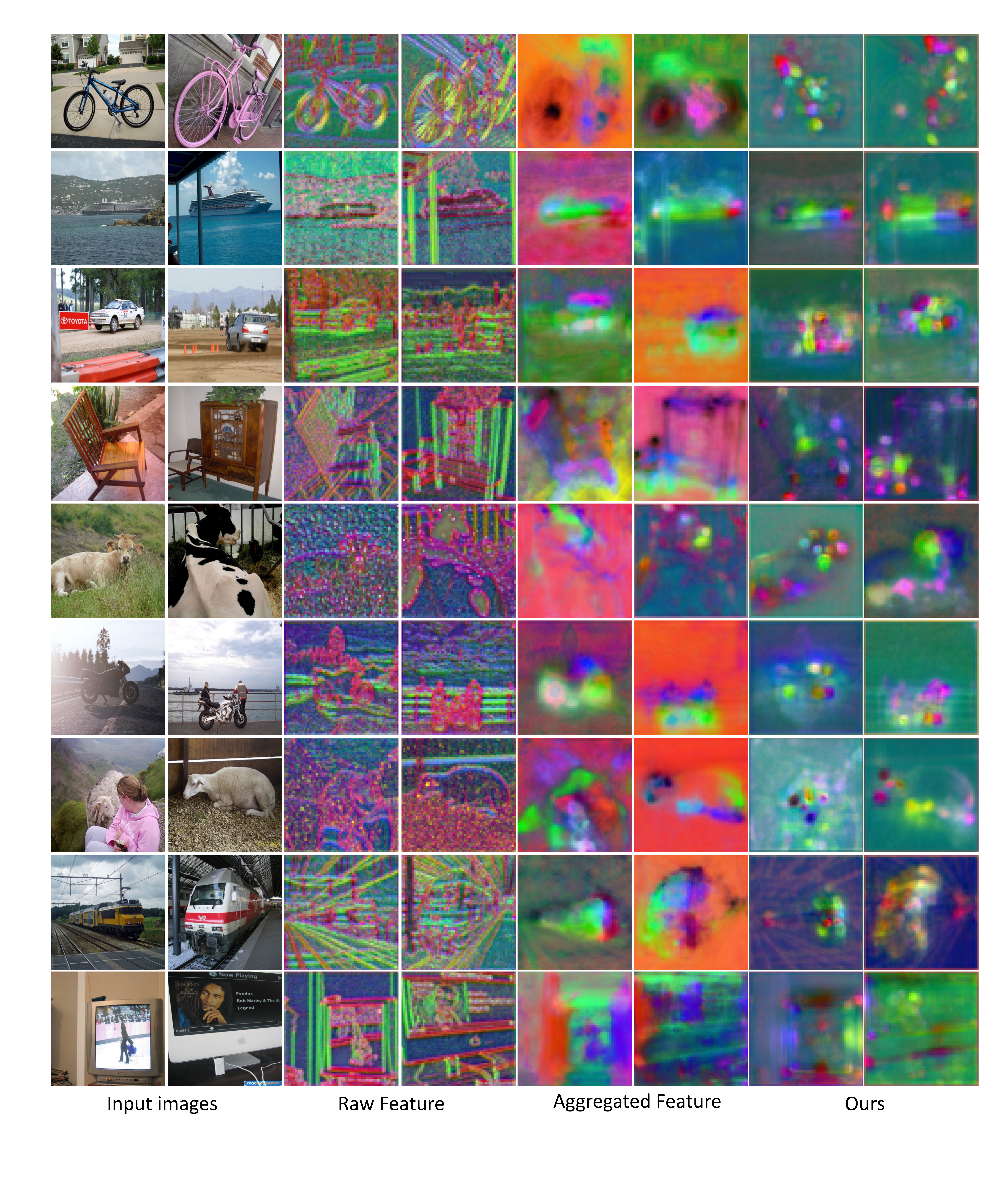}\hfill\\
    \vspace{-40pt}\caption{\textbf{Visualization of PCA results.} }
    \label{PCA1}\vspace{-10pt}
\end{figure*}

\begin{figure}
    \centering
    \includegraphics[width=0.8\linewidth]{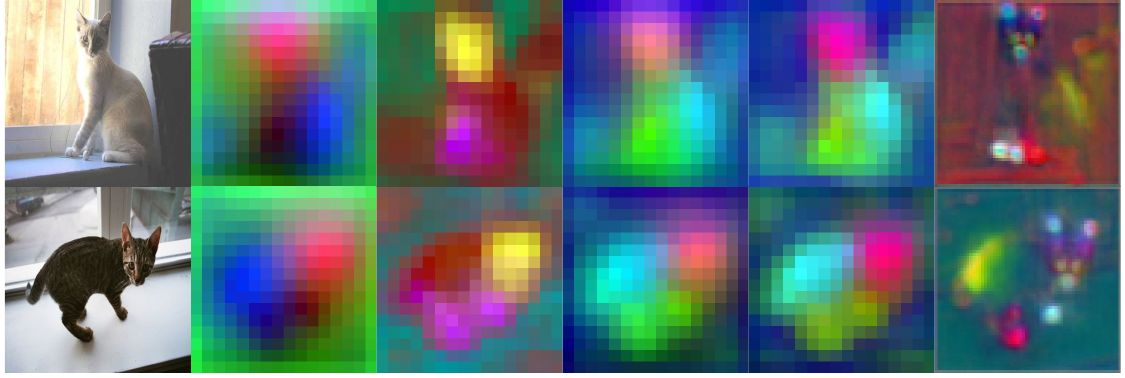}\hfill\\
    \caption{{\textbf{Qualitative comparison of ablation studies in PCA Visualizaitons. } From left to right, the PCA visualizations of input images and those of (I-V) in the component ablation table are shown.  }  }
    \label{PCA2}\vspace{-10pt}
\end{figure}

We provide more qualitative results on HPatches~\citep{balntas2017hpatches} in Fig.~\ref{hp}, ETH3D~\citep{schops2017multi} in Fig.~\ref{eth1} and Fig.~\ref{eth2} and SPair-71k~\citep{min2019spair} in Fig.~\ref{spair1} and Fig.~\ref{spair2}. We also present more visualizations of PCA and the attention maps in Fig.~\ref{attention_supp}, Fig.~\ref{PCA1} and Fig.~\ref{PCA2}.

\section{Future Works}\label{H}

In this work, we explored distinctive characteristics of both feature and cost aggregations with Transformers. From the findings, we proposed a simple yet effective architecture that benefits from their synergy. However, more advanced techniques can be incorporated to further boost the performance. For example, as future work, we believe incorporating local-correlation maps to represent higher resolution cost volume would further improve the efficiency and performance improvements can be expected if $l$ is allowed to be increased given cheaper costs to represent cost volumes. However, a simple replacement of all the global correlations within the current architecture with that of local inevitably risks losing some information, which may degrade the performance. This means that a careful design would be necessary to achieve both high efficiency and performance. Another interesting extension is that as our model currently does not explicitly model matchability or uncertainty, it may have some disadvantages when handling occlusions. To compensate,  we could design to output pixel-wise matchability scores and incorporate them into our framework, which we leave as future work.

% \begin{figure*}
%     \centering
%     \includegraphics[width=1.0\linewidth]{figure/supple_qual/cotr-ifcat_qual.pdf}\hfill\\
%     \caption{\textbf{Sparse, quasi-dense and dense comparisons.} For the sparse correspondence setting (COTR), only the confident correspondences, which are indicated as green dots, are used for evaluation. For quasi-dense correspondences (COTR+interp), the red regions are discarded at evaluation. Note that we visualize the warped image using the quasi-dense flow by COTR+interp, and then  indicate the sparse points and discarded regions as green dots and red regions, respectively, on top of the warped images. For dense evaluation (UFC), every pixel correspondence is used for evaluation, and we highlight with a white box that our method successfully finds accurate dense correspondences.   }
%     \label{cotr1}\vspace{-10pt}
% \end{figure*}

\clearpage

\begin{figure}
    \centering
    \includegraphics[width=0.8\linewidth]{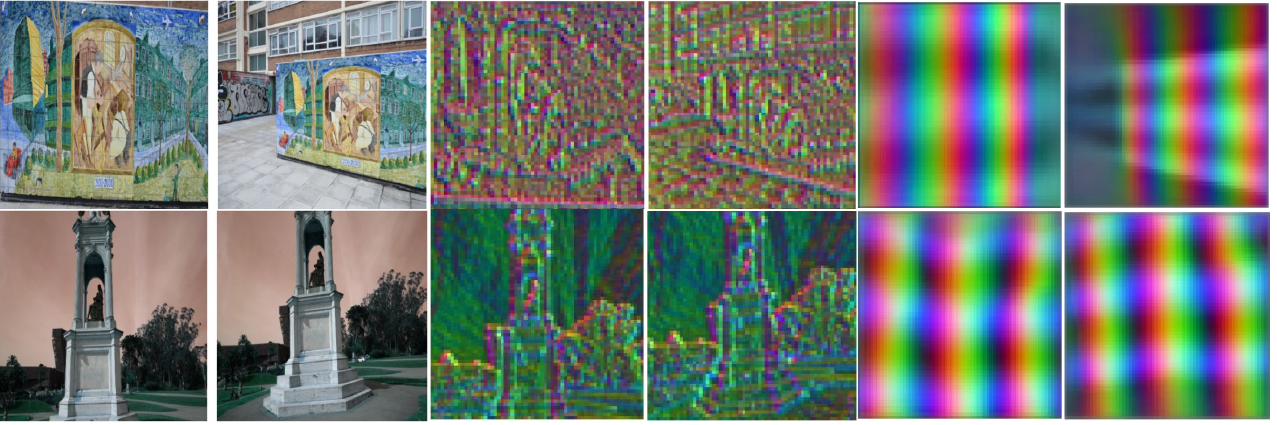}\hfill\\
    \caption{{\textbf{PCA visualizations for geometric matching. } From left to right, source, target, raw features and the features after aggregations.  }  }
    \label{PCA3}\vspace{-10pt}
\end{figure}

\clearpage
\bibliography{iclr2024_conference}
\bibliographystyle{iclr2024_conference}
\end{document}